\documentclass[lettersize,journal]{IEEEtran}
\usepackage{caption}
\usepackage{mathtools}
\usepackage{amsthm,amsmath,amsfonts}
\usepackage{algorithm}
\usepackage{algpseudocode}
\theoremstyle{definition}

\usepackage{array}
\usepackage{textcomp}
\usepackage{stfloats}
\usepackage{url}
\usepackage{verbatim}
\usepackage{graphicx}
\usepackage{subcaption}
\usepackage{booktabs}
\usepackage{multirow}
\usepackage[table]{xcolor}
\usepackage[capitalise]{cleveref}
\definecolor{cvprblue}{rgb}{0.21,0.49,0.74}
\definecolor{myorange}{rgb}{0.93,0.49,0.19}
\definecolor{myred}{rgb}{0.84,0.15,0.15}
\definecolor{mygray}{gray}{0.9}

\crefname{equation}{Eq.}{Eqs.}   
\Crefname{equation}{Eq.}{Eqs.}   
\crefname{figure}{Fig.}{Figs.}   
\Crefname{figure}{Fig.}{Figs.}   

\begin{document}

\title{Safe and Reliable Diffusion Models via Subspace Projection}

\author{Huiqiang~Chen,~~
        Tianqing~Zhu*,~~\IEEEmembership{Member,~IEEE}
        Linlin~Wang,~~
        Xin~Yu,~~
        Longxiang~Gao,~~\IEEEmembership{Senior Member,~IEEE}
        Wanlei~Zhou~~\IEEEmembership{Fellow,~IEEE}
    \thanks{*corresponding author. Huiqiang Chen, Linlin Wang, Tianqing Zhu, and Wanlei Zhou are with the Faculty of Data Science, City University of Macau, Macao, China (e-mail:cs.hqchen@gmail.com; d22092100126@cityu.edu.mo; tqzhu@cityu.edu.mo; wlzhou@cityu.edu.mo).}
    \thanks{Xin Yu is with the School of Computer Science, University of Queensland, QLD, Australia (e-mail: xin.yu@uq.edu.au).}
    \thanks{Longxiang Gao is with the Key Laboratory of Computing Power Network and Information Security, Ministry of Education, Shandong Computer Science Center, Qilu University of Technology (Shandong Academy of Sciences), Jinan, China, and also with the Shandong Provincial Key Laboratory of Computing Power Internet and Service Computing, Shandong Fundamental Research Center for Computer Science, Jinan, China (e-mail: gaolx@sdas.org)}
}
\maketitle

\begin{abstract}
Large-scale text-to-image (T2I) diffusion models have revolutionized image generation, enabling the synthesis of highly detailed visuals from textual descriptions. However, these models may inadvertently generate inappropriate content, such as copyrighted works or offensive images. While existing methods attempt to eliminate specific unwanted concepts, they often fail to ensure complete removal—allowing the concept to reappear in subtle forms. For instance, a model may successfully avoid generating images in Van Gogh’s style when explicitly prompted with ``Van Gogh", yet still reproduce his signature artwork when given the prompt ``Starry Night". In this paper, we propose SAFER, a novel and efficient approach for thoroughly removing target concepts from diffusion models. At a high level, SAFER is inspired by the observed low-dimensional structure of the text embedding space. The method first identifies a concept-specific subspace $\mathcal{S}_c$ associated with the target concept $c$. It then projects the prompt embeddings onto the complementary subspace of $\mathcal{S}_c$, effectively erasing the concept from the generated images. Since concepts can be abstract and difficult to fully capture using natural language alone, we employ textual inversion to learn an optimized embedding of the target concept from a reference image. This enables more precise subspace estimation and enhances removal performance. Furthermore, we introduce a subspace expansion strategy to ensure comprehensive and robust concept erasure. Extensive experiments demonstrate that SAFER consistently and effectively erases unwanted concepts from diffusion models while preserving generation quality.
\end{abstract}


\section{Introduction}\label{sec:intro}
\IEEEPARstart{D}{}iffusion-based text-to-image (T2I) generative models have achieved remarkable success, enabling the creation of high-quality images from simple text inputs. These models can generate a diverse range of objects, styles, and scenes with remarkable realism and variety. The success of diffusion models is largely attributed to the extensive and massive-scale data used for training. However, these datasets, which are often scraped from the web without curation, may include images that are offensive, unsafe, or protected by intellectual property. As a result, diffusion models can inadvertently learn and reproduce inappropriate \textit{concepts}, such as copyrighted artistic style, biased, or potentially harmful content~\cite{qu2023unsafe, hunter2023ai}. The open-source release of Stable Diffusion (SD)~\cite{Rombach_2022_CVPR} has made advanced image generation technology widely accessible. This highlights the urgent need to prevent these models from generating sensitive or harmful content. This raises a critical question:

\textit{How can we prevent diffusion models from generating inappropriate images? In other words, how can we erase specific concepts from diffusion models?}

\begin{figure}[t]
    \centering
    \includegraphics[width=\linewidth]{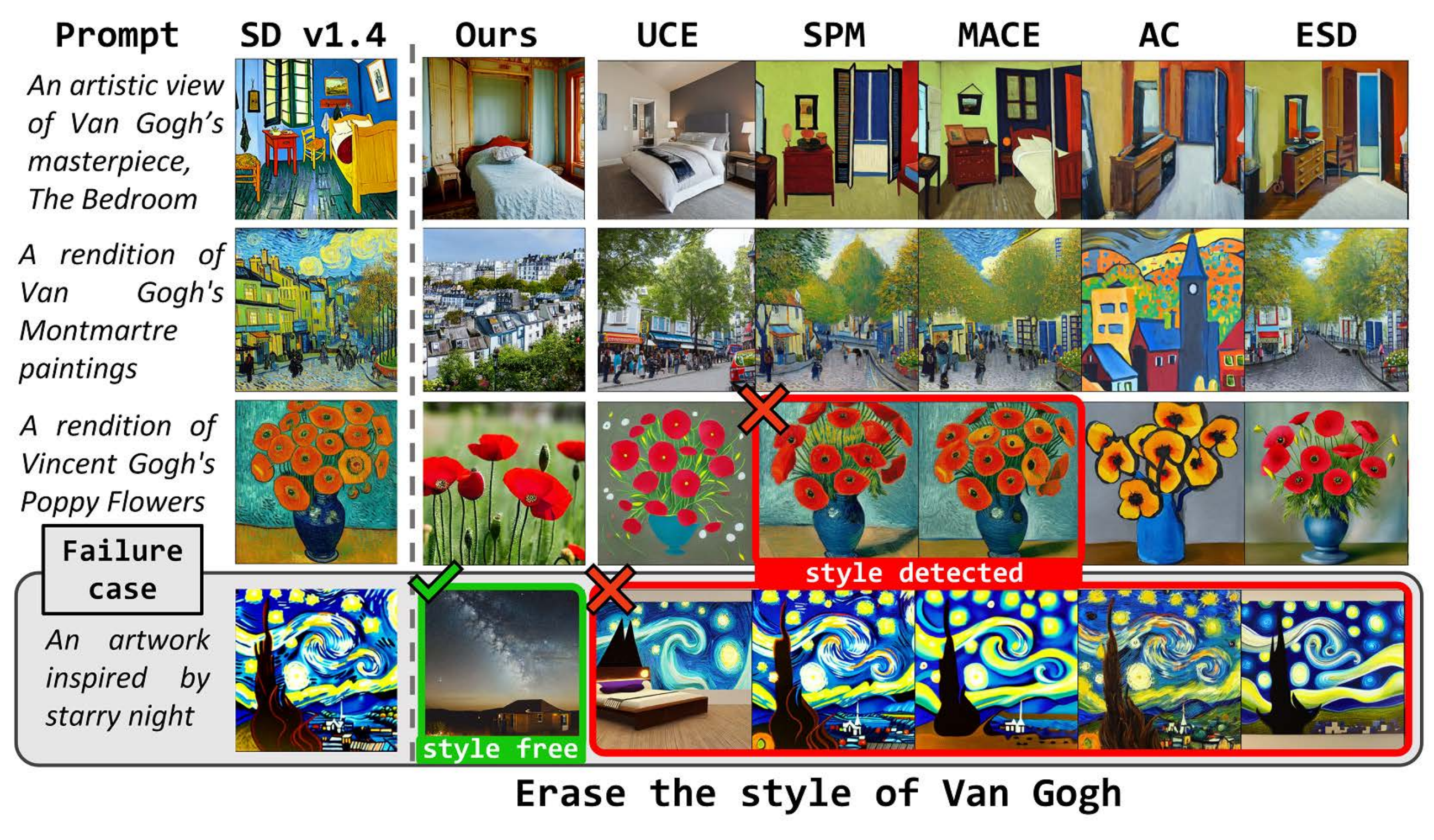}
   \caption{\textbf{Results on removing Van Gogh's style}. Existing erasing methods remain incomplete—when replacing ``\texttt{Van Gogh}" with ``\texttt{Vincent Gogh}" or prompting ``\texttt{Starry Night}", the generated images still retain Van Gogh's style. In contrast, our approach effectively removes the target style by projecting the prompt's text embedding onto the complementary subspace of the concept embedding}
    \label{fig: style_compare}
\end{figure}

Given that diffusion models may memorize inappropriate concepts from training data, machine unlearning~\cite{9844865,10607903} seems like a plausible solution. Unfortunately, as pointed out by~\cite{esd}, unlearning is not well-suited for this task. Because: 1) \textbf{It is difficult to associate a concept with a specific set of samples.} Unlearning algorithms aim to remove the influence of certain training samples as if they have never been used during training. However, the target concept is often abstract, such as a certain artistic style or nudity, which is not tied to an identifiable group of training images; More importantly, 2) \textbf{removing a subset of training samples does not necessarily eliminate the learned concept from the model.} For instance, even though Stable Diffusion v2.1~\cite{Rombach_2022_CVPR} was trained on a curated dataset with explicit content filtered out, the model still generates inappropriate images, demonstrating that filtering training data alone is insufficient.

If unlearning is not the right approach, what alternatives exist? Recent efforts have focused on implementing safety mechanisms to prevent the generation of a certain concept. These approaches can be broadly divided into two classes. The first class preserves the original model weights and implements safety measures, such as censoring model outputs through safety checkers that detect inappropriate visual content~\cite{rando2022red} or applying classifier-free guidance to steer generation away from problematic concepts~\cite{sld}. The second category modifies model weights through fine-tuning to erase specific target concepts. For example,~\cite{spm} and~\cite{lu2024mace} train multiple LoRAs~\cite{lora} for individual concepts, while~\cite{esd} fine-tunes the cross-attention module to achieve concept erasure.

While effective to some extent, these methods have drawbacks, including additional computational overhead. More critically, they often result in incomplete erasure, as they tend to block specific keywords rather than fully remove the target concept. As a result, when prompted with related words or synonyms, the erased concepts can still reappear in the generated images. This limitation underscores the need for a more robust approach that comprehensively removes concepts rather than just blocking keywords, preventing their reappearance through indirect prompts.

To address these challenges, we introduce SAFER, a novel approach to erasing concepts from diffusion models by leveraging concept subspace in the text embedding space. Our method utilizes textual inversion to capture target concepts from reference images, allowing for precise identification of the corresponding concept subspace. Furthermore, we expand this subspace to encompass a broader and more nuanced representation of the concept, ensuring complete and robust removal. The contribution of this paper is summarized as
\begin{itemize}
    \item We introduce a novel concept subspace identification method within the text embedding space, enabling precise targeting of complex concepts.
    \item We propose a subspace expansion strategy to improve the completeness of concept removal, ensuring robust erasure.
    \item We conduct extensive experiments demonstrating the superiority of our method in removing artistic styles, objects, and explicit content, as well as its capability to erase multiple concepts simultaneously.
\end{itemize}

\section{Related Works}
\label{sec: rela}
Recent research~\cite{esd,rando2022red,sld,uce} aims to mitigate the risks of harmful image generation and copyright issues associated with diffusion models. Most of the work primarily adopts two categories of methods: filtering-based approaches~\cite{rando2022red,sld,qu2023unsafe,das2024espresso} and fine-tuning~\cite{esd,spm,lu2024mace,uce,basu2023localizing,basu2024mechanistic}.

\textbf{Filtering.} The first category employs filtering methods to remove undesirable concepts. These approaches are broadly divided into in-processing and post-processing filtering. 

\textit{In-processing filtering}. In-processing filtering focuses on mitigating harmful content during the generation process. For instance, SLD~\cite{sld} leverages classifier-free guidance to suppress inappropriate elements, utilizing the model’s internal knowledge to identify and reduce harmful content during generation. ESPRESSO~\cite{das2024espresso} intervenes at the latent space level, filtering harmful content before it is fully generated. This integration makes it harder for adversarial users to bypass safeguards. However, the approach’s dependency on the model’s inherent biases and training data can limit its ability to handle harmful concepts not well-represented in the training set. Filtering methods are fundamentally reactive. Their effectiveness depends on predefined or learned concepts of harm, making them vulnerable to evasion tactics such as minor prompt adjustments. Moreover, they do not alter the underlying generative model, which retains the capacity to produce unsafe content. 

\textit{Post-processing filtering}. Post-processing techniques are applied after the generation process, blocking unsafe material to ensure only appropriate images are displayed. For example, ~\cite{rando2022red} compares the generated images against predefined sensitive concepts and blocks unsafe content. This method is straightforward and easy to implement, but its effectiveness heavily relies on the comprehensiveness of the predefined sensitive concepts. ~\cite{qu2023unsafe} deploys a fine-tuned classifier to recognize and filter harmful content categories, effectively removing various types of unsafe content. This approach is more adaptive than static concept lists, but its effectiveness is constrained by the quality of its training data. Post-process filtering operates as an external mechanism, leaving the core generative model unchanged. The model retains its capacity to produce unsafe content, and the filter’s success relies on consistently detecting and blocking harmful outputs. 

\textbf{Fine-tuning.} Fine-tuning methods directly modify model components to remove concepts, offering a proactive approach compared to filtering. These approaches can target the model at the full-scale or component-specific level.

\textit{Full model fine-tuning}. This approach involves fine-tuning the entire U-net of the diffusion model. For instance, AC~\cite{ac} fine-tunes the U-net to align predictions between prompts containing the target and corresponding anchor concepts. The goal is to redirect the model's predictions towards neutral, non-harmful outputs. FMN~\cite{fmn} introduces two losses, Attention Re-steering Loss and Visual Denoising Loss, which guide the model to suppress harmful content while maintaining generative quality. SPM~\cite{spm} attaches adapters to each U-net layer and fine-tunes these adapters to effectively erase harmful concepts. This allows for efficient fine-tuning while preserving most of the original model's structure. 

\textit{Targeted Fine-tuning}. Methods in this line focus on specific parts of the U-net. ESD~\cite{esd} investigated fine-tunes the cross-attention module x) and the rest non-cross-attention modules (ESD-u). ESD-x is more effective when the harmful concept is explicitly mentioned in the prompt, while ESD-u is better for mitigating unintended associations with implicit concepts. MACE~\cite{lu2024mace} refines the cross-attention modules using LoRA~\cite{lora} fine-tuning, extending the model's erasure capacity to handle up to 100 harmful concepts simultaneously. DiffQuickFix~\cite{basu2023localizing} and LocoEdit~\cite{basu2024mechanistic} utilize adversarial training and optimization techniques to identify and modify key model parameters. These methods focus on precisely locating the parts of the model responsible for generating harmful content, allowing for more focused removal without significantly affecting unrelated model capabilities. 

\section{Preliminary}
\subsection{Diffusion models} Diffusion models are a class of generative models that generate data by progressively removing noise from intermediate representations \cite{ho2020denoising, song_improved_2020, dhariwal_diffusion_2021}. A diffusion model can be decomposed into the forward and reverse diffusion processes.

The forward diffusion process gradually corrupts a data sample $x_0$ drawn from the real data distribution $q(x_0)$ (e.g., an image) by adding Gaussian noise step by step. This process can be modeled as a fixed Markov chain spanning $T$ steps. The transition from $x_{t-1}$ to $x_t$ is defined as
\begin{equation}
    q(x_t | x_{t-1}) = \mathcal{N}(x_t; \sqrt{1-\beta_t}x_{t-1}, \beta_t\mathbf{I}),
\end{equation}
where $\beta_t \in (0, 1)$ is the scheduled variance at step $t$. By iterating this process, we can directly express $x_t$ in terms of the original image $x_0$:
\begin{equation}
    x_t = \sqrt{\Bar{\alpha}_t}x_0 + \sqrt{1 - \Bar{\alpha}_t}\epsilon,
\end{equation}
where $\epsilon \sim \mathcal{N}(0, I)$ is standard Gaussian noise and $\Bar{\alpha}_t = \prod_{i=1}^{t}(1-\beta_{i})$. As $t$ increases, $x_t$ becomes nearly indistinguishable from pure Gaussian noise.

The reverse diffusion process is learned using a neural network, which estimates the noise at each step to progressively reconstruct the original data distribution $q(x_0)$. Starting from pure noise $x_T \sim \mathcal{N}(0, I)$, the model gradually refines the sample step by step.

At each step $t$, the denoising network $\epsilon_\theta(x_t, t)$ predicts the noise that was added to $x_0$. Using this estimated noise, we can compute an estimate of $x_0$:
\begin{equation}
    \hat{x}_0^t = \frac{x_t - \sqrt{1 - \Bar{\alpha}_t} \epsilon_{\theta}(x_t, t)}{\sqrt{\Bar{\alpha}_t}}.
    \label{equation:x_0_hat}
\end{equation}
Given $\hat{x}_0^t$, we can compute$x_{t-1}$as
\begin{equation}
    x_{t-1} = \sqrt{\Bar{\alpha}_{t-1}} \hat{x}_0^t + \sqrt{1 - \Bar{\alpha}_{t-1}} \epsilon_{\theta}(x_t, t) + \sigma_t z,
    \label{equation:x_t-1}
\end{equation}
where $z \sim \mathcal{N}(0, I)$ is sampled Gaussian noise, and $\sigma_t$ is a learned or fixed variance term that controls the stochasticity of the process.

\subsection{Text-conditional diffusion models} 
Text-conditional diffusion models, such as Stable Diffusion \cite{rombach2022high}, extend the diffusion process by adding text-based conditioning, allowing control over the generated content using natural language prompts. These models utilize classifier-free diffusion guidance \cite{rombach2022high} to influence the generation process according to the provided prompt. Given a text prompt $p$, the corresponding embedding $e_p = E_{text}(p)$ is computed using a pre-trained CLIP text encoder $E_{text}(\cdot)$ \cite{Radford2021LearningTV, cherti2023reproducible}. During the reverse diffusion process, generation follows \Cref{equation:x_0_hat,equation:x_t-1}, but the predicted noise $\epsilon_{\theta}(x_t)$ is adjusted to include the influence of the text prompt:
\begin{equation*}
    \epsilon_{\theta}(x_t, t, e_{\emptyset}) + s \left( \epsilon_{\theta}(x_t, t, e_p) - \epsilon_{\theta}(x_t, t, e_{\emptyset}) \right),
\end{equation*}
where $e_{\emptyset}$ represents the embedding of an empty prompt, and $s$ is the guidance scale, which determines how strongly the generated output is influenced by the text prompt. A higher value of $s$ results in outputs that are more closely aligned with the input prompt.

\subsection{Textual inversion} 
Text-to-image diffusion models can synthesize a variety of images conditioned on text prompts. However, it is hard to design a prompt that can guide the diffusion model to generate an object or a style in a precise way. Textual inversion \cite{gal2022image} is a technique that maps a group of images into a single pseudo-word, which can be used in prompts to generate highly personalized photos. 
Despite the success of text-to-image diffusion models, crafting prompts that yield precisely controlled outcomes remains challenging. Text prompts often need to be highly specific to generate a desired style, composition, or object, an effort known as ``prompt engineering''. Textual Inversion \cite{gal2022image} has been introduced to address this challenge by mapping visual concepts, such as an object or a style, to new pseudo-words in the model's embedding space. This process involves encoding a set of images to create a unique token that represents a specific object, style, or scene, allowing users to reuse this token in other prompts to reproduce highly personalized results \cite{gal2022image}. Formally, given an embedding space $\mathcal{E}$ and a set of images $\{I_i\}$ representing a target concept, textual inversion learns an optimized embedding $v^*$ such that:
\begin{equation}
    v^* = \arg \min_v \sum_i \mathcal{L}(M(v, t), I_i),
\end{equation}
where $M$ represents the diffusion model, $t$ is a textual prompt, and $\mathcal{L}$ is the reconstruction loss between the generated and target images. 

\begin{figure*}[t]
    \centering
    \includegraphics[width=\linewidth]{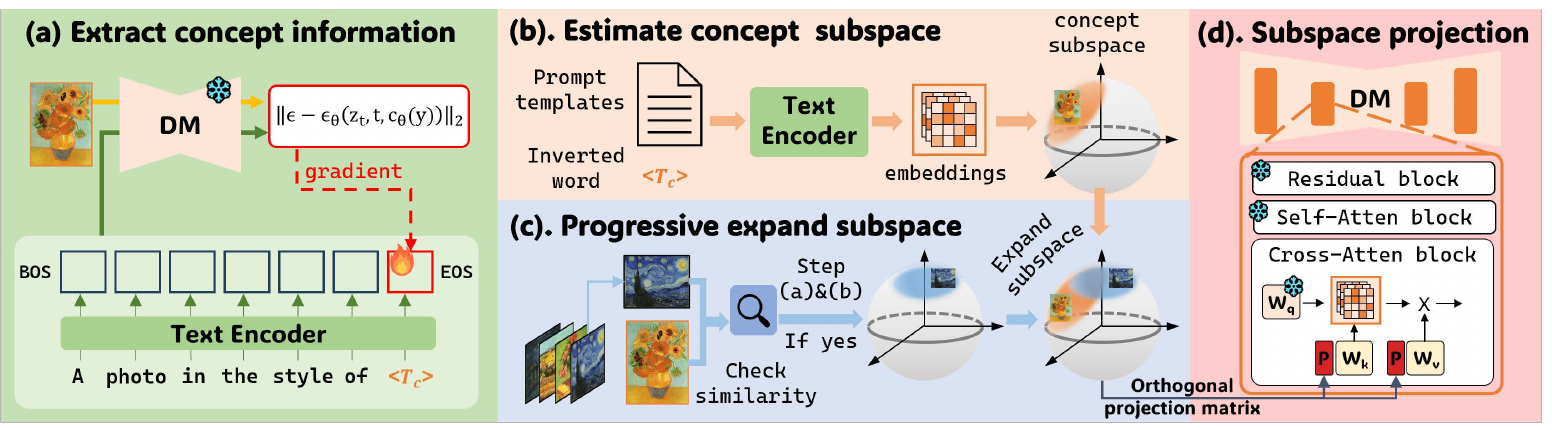}
    \caption{\textbf{Overview of the Proposed Framework}. (a) Starting with a reference image containing the target concept, we invert its visual characteristics into a specialized token $\mathcal{T}_c$ paired with an optimized embedding within the text encoder's vocabulary. (b) The concept subspace is estimated in the text embedding space by generating prompts that combine prompt templates with the specialized token derived from the reference image. The first principal component of these prompt embeddings forms the basis of the target concept subspace. (c) The concept subspace is progressively expanded to approximate the target concept more comprehensively, ensuring a thorough erasure. (d) After identifying the target concept subspace, the text embedding is projected onto its complementary subspace. The resulting projection matrix is integrated into the weights of the cross-attention layers.}
    \label{fig: pipe}
\end{figure*}

\section{Method}
In this section, we introduce our approach for identifying the concept subspace in diffusion models. First, we define the concept subspace and establish its structure through a set of diverse prompts. Next, we enhance this identification using textual inversion to extract concepts directly from reference images. Finally, we propose a method for progressively expanding the subspace to ensure comprehensive concept removal.

\subsection{Subspace in Diffusion Model} \label{sec: sub struct}
To identify the concept subspace $\mathcal{S}_c$ in a diffusion model, we construct a set of prompts $\mathcal{P}^c=\{p^c_i\}_{i=1}^N$ that explicitly reference the target concept $c$, such as Van Gogh's style. These prompts are generated using ChatGPT~\cite{gpt} and designed to be diverse, ensuring that each prompt is unique except for the inclusion of concept $c$. Examples include \texttt{A } $\langle obj \rangle$ \texttt{ in }$\langle c \rangle$\texttt{'s style} and \texttt{A painting of }$\langle obj\rangle$\texttt{ inspired by the style of} $\langle c\rangle$.

Each prompt $p^c_i$ is processed through the text encoder of the diffusion model, producing a corresponding text embedding $e^c_i \in \mathbf{R}^d$. We model these embeddings as

\begin{equation}
    e^{c}_i = \alpha_i v_{c} + o_i + \kappa_i,
\end{equation}
where $o_i \in \mathbf{R}^d$ is an object-related component, which is independently sampled since the $\langle obj \rangle$ is randomly chosen when constructing the prompt list. The vector $v_c \in \mathbf{R}^d$ captures the common component shared across prompts, associated with the concept $c$. The scalar $\alpha_i \in \mathbf{R}$ controls the influence of $v_c$ for each prompt $p^c_i$, while $\kappa_i \in \mathbf{R}^d$ represents unmodeled variations, also independently drawn across samples. To simplify notation, we assume $e^{c}_i$ is mean-centered, i.e., $\mathbb{E}[o] = 0$, $\mathbb{E}[\kappa] = 0$.

\begin{figure}[htp]
    \centering
    \begin{subfigure}[b]{0.24\textwidth}
        \includegraphics[width=\textwidth]{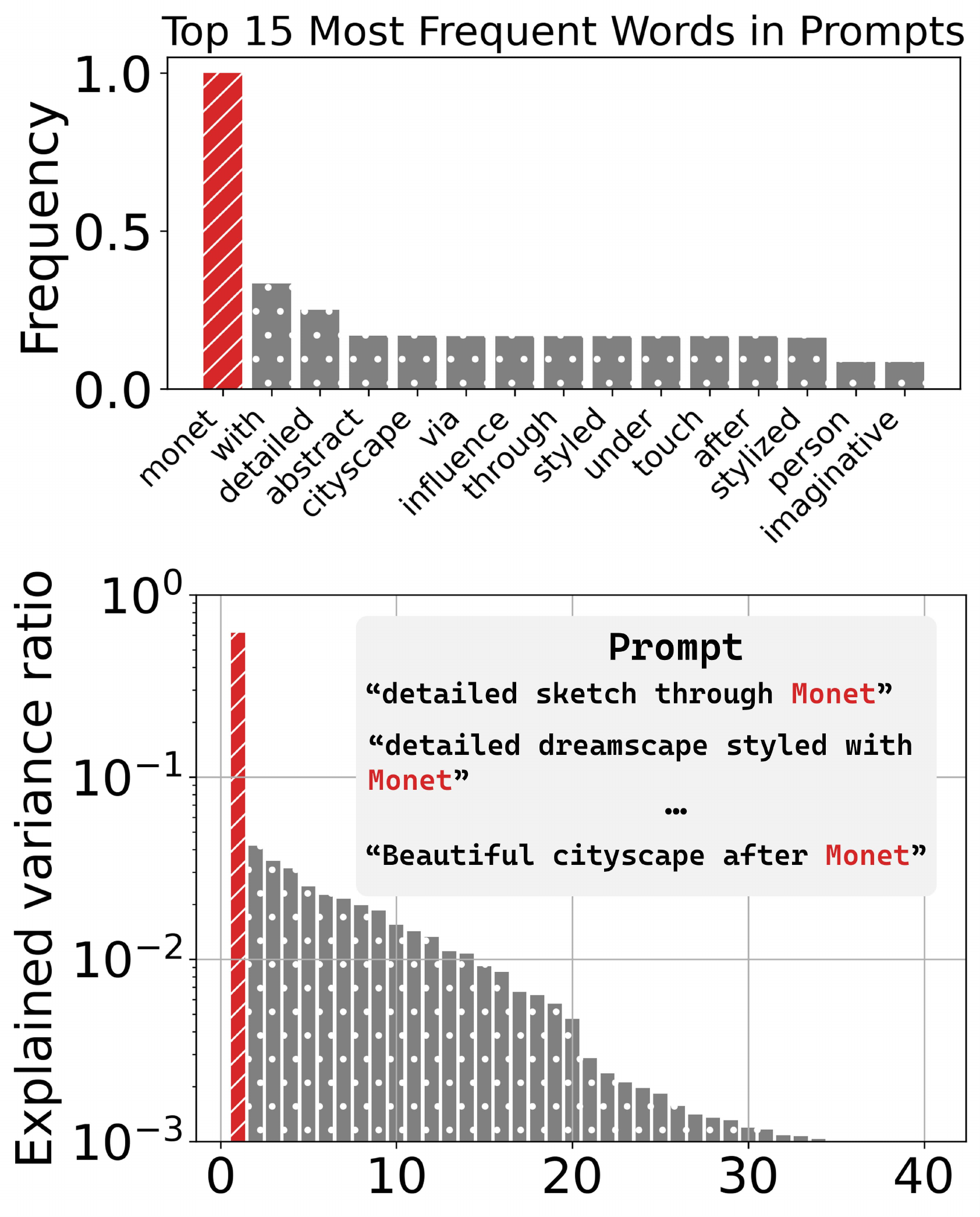}
        \caption{Style}
    \end{subfigure}
    \hfill
    \begin{subfigure}[b]{0.24\textwidth}
        \includegraphics[width=\textwidth]{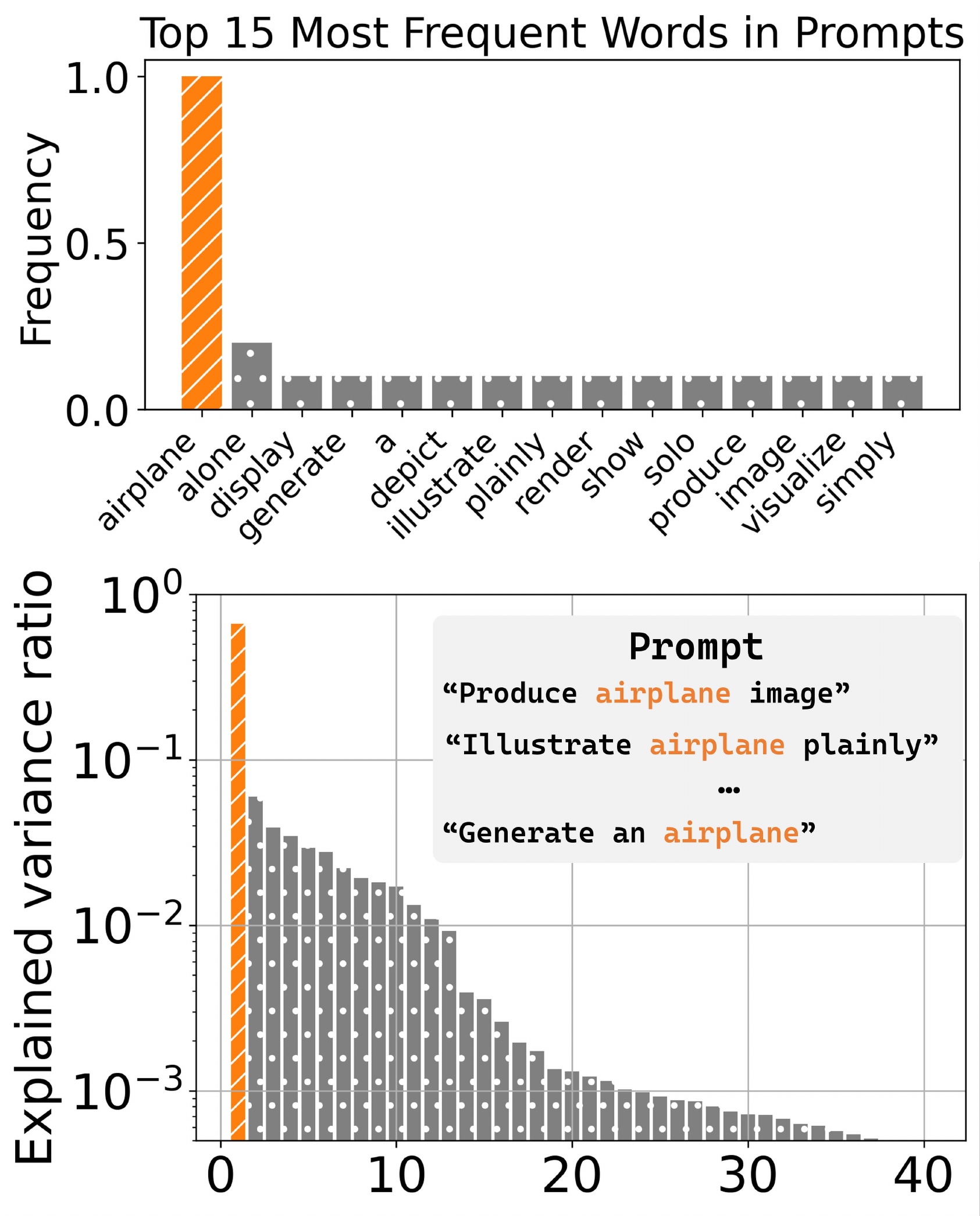}
        \caption{Object}
    \end{subfigure}
    \caption{\textbf{SVD results of text embeddings}. The explained variance of the first component is significantly higher than that of the second, indicating a strong correlation between the primary components of the signal, particularly for (a) the style of \textcolor{myred}{Monet} and (b) the object \textcolor{myorange}{Airplane}.}
    \label{fig: svd}
\end{figure}

Denote the embeddings of each prompt as rows in a matrix $\Sigma_{e^c}\in \mathbf{R}^{N\times d}$, where each row corresponds to an embedding $e^{c}_i$ for the $i$-th prompt. Thus, the matrix $\Sigma_{e^c}$ takes the form of
\begin{equation}
    \Sigma_{e^c} = \alpha v^T_c+\mathbf{O}+K.
\end{equation}
where $\alpha$ is the vector of coefficients, $\mathbf{O}$ contains object-related components, and $K$ represents the noise components. Considering the covariance matrix of the embeddings, represented by $\Sigma_{e^c}\Sigma_{e^c}^T$, and taking the expectation over the randomness in the prompts, we have:
\begin{equation}
    \mathbb{E}[\Sigma_{e^c}\Sigma_{e^c}^T]=\mathbb{E}[\alpha^2 v_cv^T_c+\mathbf{O}\mathbf{O}^T+KK^T],
\end{equation}
where the concept component term $\mathbb{E}[\alpha^2 v_cv^T_c]=\sigma_{\alpha}^2v_cv^T_c$ has rank $1$ and aligns with $v_c$. For the object and noise terms, we have $\mathbb{E}[\mathbf{O}\mathbf{O}^T]=0$  due to the independence of $o_i$, and similarly, $\mathbb{E}[KK^T]=0$. This implies that $\Sigma_{e^c}$ approximately lies in a subspace spanned by $v_{c}$. 

To verify this, we conduct two experiments. In the first, we set the concept $c$ as a style, such as Monet's style, and in the second, we set $c$ as airplane. In each experiment, we generate a list of prompts involving $c$ with varying content. These prompts were processed by the text encoder to obtain the embeddings $\Sigma_{e^c}$. We then perform SVD on $\Sigma_{e^c}$ and examine the explained variance ratio(\Cref{fig: svd}). The results confirm that the first principal component captures the majority of the variance, supporting our hypothesis that the target concept is well-represented within this subspace.

\subsection{Identify Concept Subspace via Textual Inversion} \label{sec: ti}
\begin{algorithm}[t]
    \caption{Identify concept subspace via textual inversion}
    \label{alg: tis}
    \textbf{Input}:An image $x_i$; text encoder of diffusion model $f$; a set of prompt template $\mathcal{P}^c$. \\   
    \textbf{Output}: basis of concept subspace $U_1$.
    \begin{algorithmic}[1]
        \State $\mathcal{T}_c \gets$ \textit{Textual inversion}$(x_i, \theta)$
        \For{$ p_i \in \mathcal{P}^c$}
            \State $e_i\gets$ $f$(\textit{Combine}($p_i$, $\mathcal{T}_c$)) 
        \EndFor
        \State $\mathbf{U}, \Sigma, \mathbf{V} \gets \textit{SVD}(\textit{Concatenate($e_1, e_2,...,e_n$)})$
        \State  \Return $U_1^i=\mathbf{U}[1]$ as the basis of concept subspace.
    \end{algorithmic}
\end{algorithm}

Accurately estimating the concept subspace is crucial for effectively removing the concept. However, using natural language prompts to identify a concept subspace, as introduced in the previous section, may be insufficient because language alone cannot fully capture certain concepts. For instance, styles can be subjective and complex, encompassing subtle attributes like mood, texture, and overall aesthetic coherence, which are challenging to describe in words. Similarly, objects can be defined in multiple ways, including synonyms or implicit references. Therefore, even well-crafted prompts may fail to precisely convey the intended concept.

A more effective approach is to use an image to represent the concept, such as a visual example of a particular concept. We adopt the technique of textual inversion~\cite{gal2022image} to extract the concept from a reference image, translating its visual attributes into an internalized, manipulable representation within the model. This approach encodes intricate details and stylistic nuances, providing a more reliable basis for identifying concept subspaces. Given a reference image $x_{ref}$ that contains the target concept $c$, we modify the text encoding process to add a new ``word", denoted as $\mathcal{T}_c$, paired with a new embedding $\textbf{u}_c$ in the text encoder's vocabulary. The optimization goal is defined as
\begin{equation}
    \textbf{u}_* = \arg\min_{\textbf{u}} \mathbb{E}_{z\sim \varepsilon(x_{ref}),y,\epsilon\sim \mathcal{N}(0,I),t}\left[\left\|\epsilon-\epsilon_{\theta}(z_t,t,c_{\theta}(y))\right\|\right],
\end{equation}
where $y$ is a prompt that contains the inverted word $\mathcal{T}_c$. After optimization, $\mathcal{T}_{c}$ can be incorporated into prompts to guide the diffusion model to generate images that include the target concept $c$ as represented by $x_{ref}$. We then feed a list of such prompts into the model, collect the corresponding text embeddings $\Sigma_{e^c} = \{e^c_0, e^c_1, \ldots, e^c_n\}$, and perform Singular Value Decomposition (SVD) to extract the dominant component as follows:

\begin{equation} \label{eq: svd}
    \mathbf{U},\mathbf{\Sigma}, \mathbf{V} = \textbf{SVD}(\Sigma_{e^c}).
\end{equation}
The first left singular vector in the $\mathbf{U}$, denoted as $U_1$, is selected as the basis of the concept subspace $\mathcal{S}_c$.

\subsection{Subspace Projection}
Once we have identified the subspace $\mathcal{S}_c$ corresponding to the target concept $c$, we can manipulate the embedding to either remove or amplify this concept. We project the original embedding out of the subspace $\mathcal{S}_c$ to remove the concept. The projection matrix is defined as

\begin{equation}\label{eq: remove_proj}
    \mathbf{P} = \mathbf{I} - U_1 U_1^T,
\end{equation}
where $\mathbf{I}$ is the identity matrix. This operation effectively removes the component of the embedding that lies within the concept subspace, ensuring that the modified embedding no longer contains the target concept $c$. Prompting the model with projected embedding ensures the generated images contain no trace of the target concept.

Alternatively, we can add a new component sampled from the subspace $\mathcal{S}_c$ to the original embedding to introduce the target concept in the generated image. The projection matrix for adding the concept is given by:
\begin{equation}\label{eq: add_proj}
    \mathbf{P} = \mathbf{I} + \lambda U_1 U_1^T,
\end{equation}
$\lambda$ is a scaling coefficient that controls the intensity of the added concept. By augmenting the original embedding with the target concept, the modified embedding now contains the desired concept. When the diffusion model processes this augmented embedding, it will generate images that reflect the presence of the corresponding concept.

\begin{figure}[t]
    \centering
    \includegraphics[width=\linewidth]{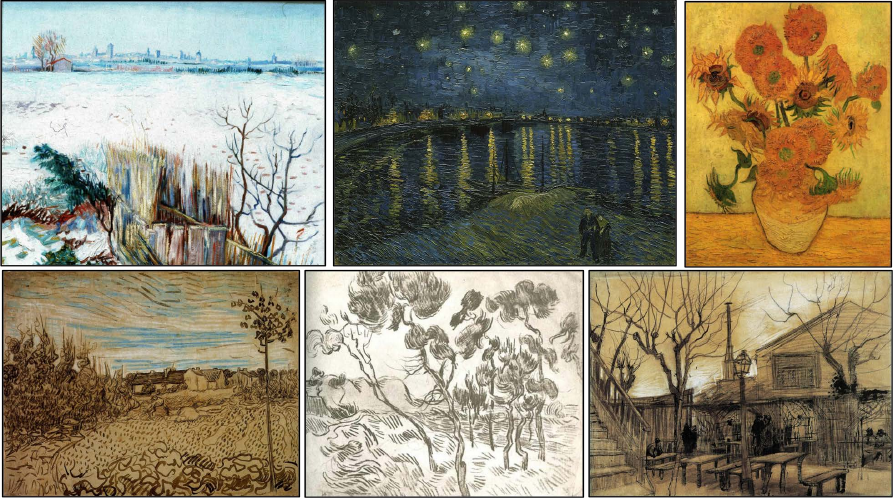}
    \caption{\textbf{Paintings of Van Gogh in different styles}. Using one image or a unified style to describe them all is hard. As such, we propose to progressively expand the subspace to include different styles of Van Gogh.}
    \label{fig:vg-paints}
\end{figure}

\subsection{Progressively expand concept subspace}\label{sec: pes}
\begin{algorithm}[t]
    \caption{Progressively Expand Subspace}
    \label{alg: pes}
    \textbf{Input}: 
    Reference image(s) $X_{\text{ref}}$ containing concept $c$; Expansion threshold $\tau$. \\
    \textbf{Output}: Projection matrix $\mathbf{P}$ for concept removal/amplification.
    \begin{algorithmic}[1]
        \State Select one image $x_i$ from $X_{\text{ref}}$ and get the initial concept subspace $U_1^i$ via~\Cref{alg: tis}      
        \State Get projection matrix $\mathbf{P} = (\mathbf{I} - U_1^i U_1^{iT})$.
        \Statex \textit{Progressively Expand Concept Subspace}
        \For{each new reference image $x_j \in X_{\text{ref}}$}
            \State Compute embedding $f_{\text{ViT}}(x_j)$ using a ViT model.
            \State Compute cosine similarity score:
            \[
            \text{sim}(x_i, x_j) = \cos\left(f_{\text{ViT}}(x_i), f_{\text{ViT}}(x_j)\right)
            \]
            \If{$\text{sim}(x_i, x_j) > \tau$}
                \State Compute SVD on new embeddings to obtain $U_1^j$.
                \State Update projection matrix:
                \[
                \mathbf{P} = \mathbf{P}(\mathbf{I} - U_1^j U_1^{jT})
                \]
            \EndIf
        \EndFor
        \State \Return $\mathbf{P}$
    \end{algorithmic}
\end{algorithm}

Using a fixed subspace is insufficient to fully capture the target concept. For instance, let's consider removing the Van Gogh's style. As shown in \Cref{fig:vg-paints}, Van Gogh's paintings encompass various styles, indicating the need for multiple subspaces to capture his diverse styles; otherwise, erasure is incomplete. \Cref{fig: style_compare} provides such an example. Images from SPM~\cite{spm} and MACE~\cite{lu2024mace} show no Van Gogh influence when prompted with \texttt{Van Gogh}. However, when replacing \texttt{Van Gogh} with \texttt{Vincent Gogh}, the generated images show a certain level of similarity to those produced by the original SD v1.4, indicating that the erasure is incomplete. In a more severe case, where we prompt the model with \texttt{Starry Night}, an iconic painting by Van Gogh, the generated images exhibit Van Gogh's style.

An effective solution is to progressively expand the subspace to better approximate the target concept subspace. Consider the example of erasing the style of Van Gogh. We assume a list of Van Gogh's paintings, denoted as $X_{ref}=\{x_i\}_{i=1}^M$. Starting with an image $x_i\in X_{ref}$, we obtain the corresponding concept subspace basis $U_1^i$ as outlined in~\Cref{alg: tis}. For a new image $x_j\in X_{ref}$, we compare the similarity between $x_i$ and $x_j$ to decide whether to expand the subspace. Each image $x_i$ is passed to a ViT backbone and then projected to a $d$-dimensional vector $f_{\text{ViT}}(x_i)\in \mathbb{R}^d$. We compute the cosine similarity between these vectors:
\begin{equation}\label{eq: sim_score}
    sim(x_i, x_j)=\cos({f_{\text{ViT}}(x_i), f_{\text{ViT}}(x_j)}).
\end{equation}
We expand the concept subspace $\mathcal{S}_c$ to include additional information from $x_j$ whenever $sim(x_i, x_j)$ exceeds a threshold. Let $U_1^i$ and $U_1^j$ be the subspace basis for $x_i$ and $x_j$ obtained from~\Cref{alg: tis}. The updated projection matrix is formulated as follows:
\begin{equation}\label{eq: combined_remove_proj}
    \mathbf{P} = \mathbf{P}_i\mathbf{P}_j=(\mathbf{I} - U_1^i U_1^{iT})(\mathbf{I}-U_1^j U_1^{jT}).
\end{equation}
\Cref{eq: combined_remove_proj} sequentially projects out the concept encoded in $x_i$ and $x_j$. This process is iteratively repeated on new images to ensure a thorough erasure. The full procedure is detailed in~\Cref{alg: pes}. Once the projection is obtained, it is integrated into the model's weights as outlined in~\Cref{alg: mpm}.

\subsection{Discussion}
Our method identifies a concept subspace $\mathcal{S}_c$ without requiring any additional training, making it efficient and easily extendable. In contrast to existing methods such as~\cite{esd} and~\cite{spm}, which involve fine-tuning or training supplementary modules, our approach is more efficient, cost-effective, and straightforward. The process only requires accessing the text encoder to encode the prepared prompts and determine the concept subspace, which is then used to remove the concept. This process requires no additional training.

Existing methods like~\cite{spm, lu2024mace} rely on post-processing or blacklisting techniques, which are vulnerable to circumvention by removing modules, disabling filters, or using synonyms for the target concept. Our approach, however, uses textual inversion to accurately capture the concept directly from the image, thereby preventing evasion through synonyms. Moreover, the projection matrix is directly integrated into the weights of the diffusion model, specifically within the cross-attention layers, embedding the concept removal as an intrinsic part of the model. This makes it difficult to bypass, even for users with full access to the model parameters.

By embedding the projection matrix into the core layers of the diffusion model, our method ensures effective and resilient concept removal that cannot be disabled after deployment. It is also highly adaptable and capable of handling multiple concepts by expanding the concept subspace $\mathcal{S}_c$.

\begin{algorithm}[t]
    \caption{Merging projection matrix into weights}
    \label{alg: mpm}
    \textbf{Input}: Projection matrix $\textbf{P}$. \\
    \textbf{Output}: Modified diffusion model $M$.
    \begin{algorithmic}[1]
        \State Select a target layer $w_l$ in $M$ and modify its weight as 
        \[
            w_l = \mathbf{P}w_l
        \]
        \State \Return Modified diffusion model $M$
    \end{algorithmic}
\end{algorithm}

\section{Experiment}
In this section, We evaluate the proposed method for erasing various concepts, including artistic style, object, and nudity content. We compare our method against SOTA baselines including ESD~\cite{esd}, FMN~\cite{esd}, UCE~\cite{uce}, SPM~\cite{spm}, SLD~\cite{sld}, AC~\cite{ac}, and MACE~\cite{lu2024mace}. We choose SD v1.4\cite{Rombach_2022_CVPR} as the base model for its easy access, widespread use, and strong generation capabilities. 

\subsection{Existence of Concept Subspace}
We first validate the existence of a concept subspace by setting the target concept $c$ as Van Gogh's style and attempting to remove it through subspace projection. Specifically, we select different components of the $\mathbf{U}$ matrix from \Cref{eq: svd} as the basis for the concept subspace $\mathcal{S}_c$. The prompt embeddings are then projected onto the complementary subspace to remove the influence of $c$. The projection matrix is defined as $\mathbf{P} = \mathbf{I} - U_j U_j^T$, where $U_j$ represents the $j$-th principal component of $\mathbf{U}$. 

\begin{figure}[t] 
    \centering
    \includegraphics[width=\linewidth]{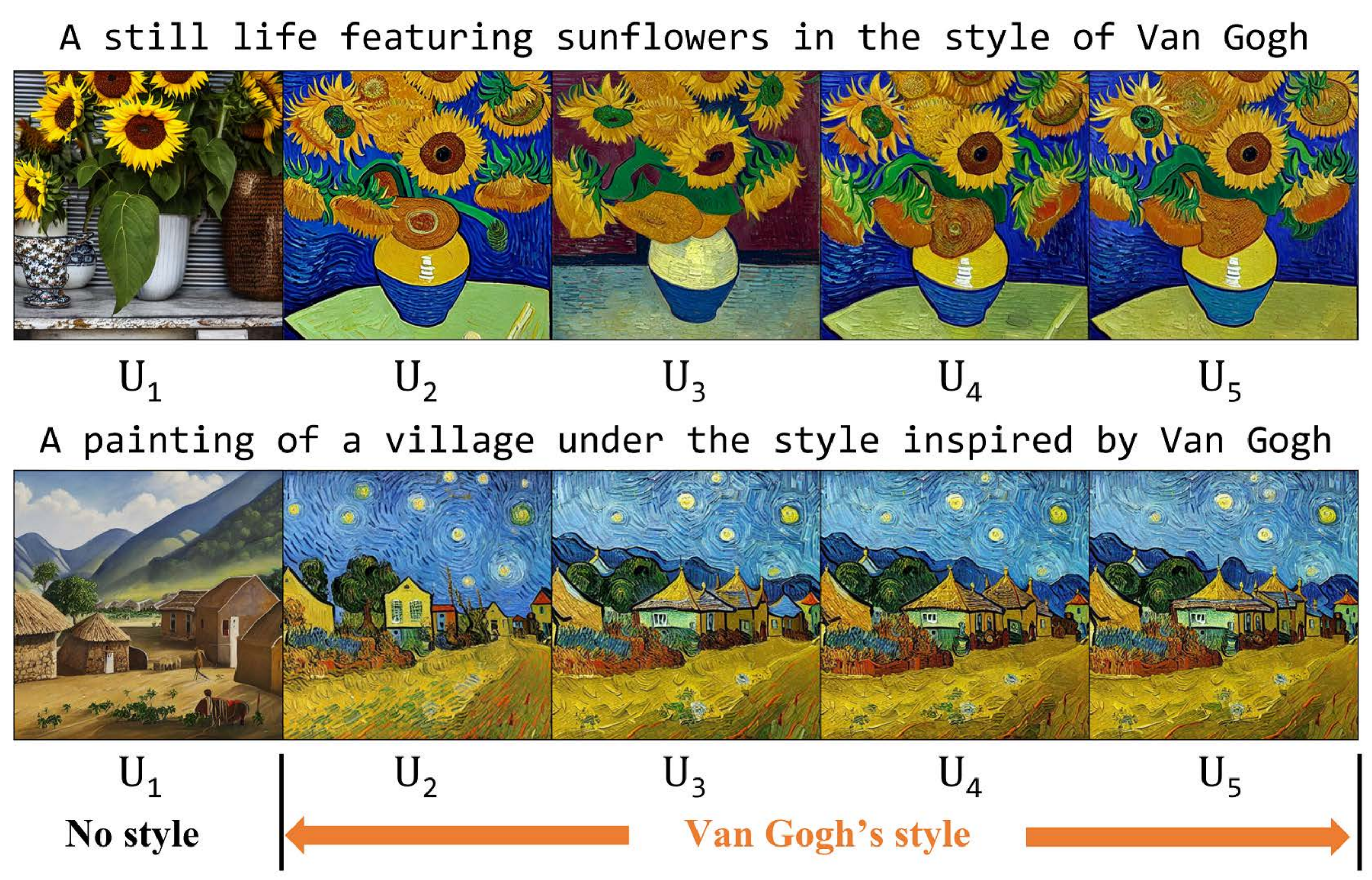}
    \caption{\textbf{Style Information Captured via a subspace.} We use different projection matrixes in the form of $I-U_jU_j^T$ in removing the Van Gogh's style. The generated images demonstrate that the style information can be captured via the subspace spanned by the first principal component of SVD results. The prompts used for image generation are displayed above each image.}
\label{fig: diff_rank}
\end{figure}

The results, shown in \Cref{fig: diff_rank}, demonstrate the effectiveness of this approach. The first column of images exhibits no trace of Van Gogh's style, whereas images in the subsequent columns retain varying degrees of stylistic influence. Notably, selecting $U_1$ as the basis for the concept subspace successfully eliminates Van Gogh’s style, whereas alternative choices, such as $U_2$, do not yield the same effect. This suggests that the primary stylistic information is captured by the first principal component, while other components encode unrelated variations or residual information. Furthermore, the projection process removes the style while preserving the primary object in the image, confirming that style and object-related features are largely separable in the embedding space. 

These findings provide strong empirical evidence for the existence of a well-defined concept subspace within the text embedding space, aligning with our theoretical analysis in \Cref{sec: sub struct}. The results further indicate that concept removal via subspace projection is effective when the subspace is accurately identified, highlighting the importance of selecting the most informative components.

\subsection{Concept Manipulation via Projection}
\begin{figure}[t]
    \centering
    \includegraphics[width=\linewidth]{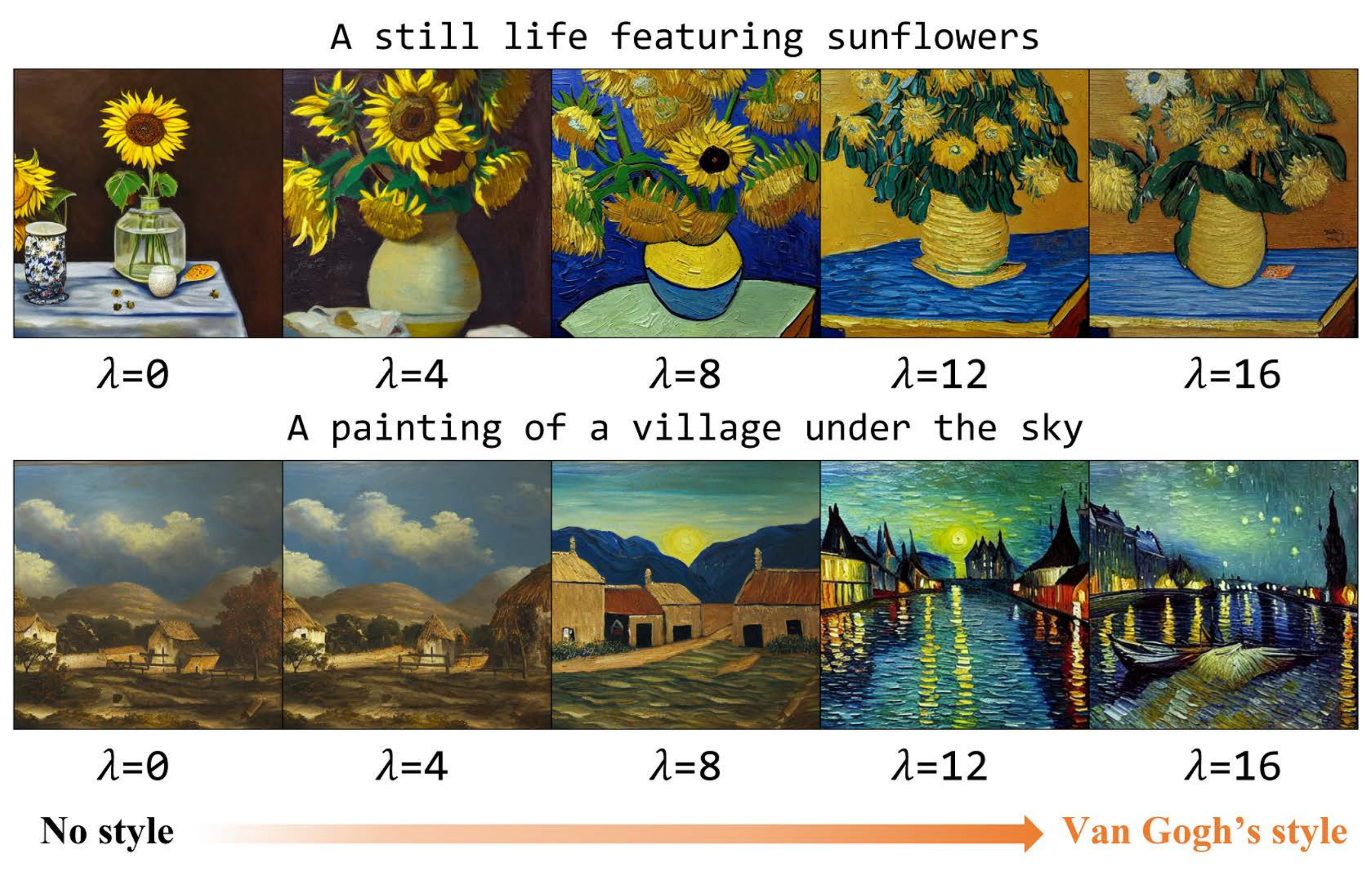}
     \caption{\textbf{Effect of varying $\lambda$ on Amplifying Van Gogh's Style.} The generated images illustrate how adjusting $\lambda$ in the projection matrix controls the incorporation of Van Gogh's style. For $\lambda = 0$, the images exhibit no stylistic influence, while increasing $\lambda$ progressively enhances the characteristic features of Van Gogh’s style. }
    \label{fig: diff_lam}
\end{figure}

In addition to removing the target concept $c$, we can also inject the concept $c$ into the generated image by amplifying the projection of $E_{text}(\text{prompt})$ onto the subspace $\mathcal{S}_c$. This is achieved using the projection transformation defined in~\Cref{eq: add_proj}. In our example, we set $c$ as Van Gogh's style and use $U_1$, obtained in~\Cref{alg: tis}, as the basis for the concept subspace.

To evaluate this approach, we prepare two prompts to generate a painting of a sunflower and a village without specifying any artistic style. We then vary the scaling factor $\lambda$ in~\Cref{eq: add_proj} to generate five images for each prompt while maintaining a fixed random seed. The results are presented in~\Cref{fig: diff_lam}. In the first column, where $\lambda = 0$, we apply an identical projection matrix, resulting in images that do not exhibit Van Gogh’s style. As $\lambda$ increases, the style of the generated images gradually shifts, progressively incorporating Van Gogh's distinctive artistic characteristics, such as swirling brushstrokes and bold color contrasts.

These results demonstrate that by adjusting $\lambda$, the contribution of the concept subspace can be finely controlled, allowing for a gradual and precise interpolation between neutral and highly stylized outputs. A smaller $\lambda$ introduces only subtle stylistic elements, while a larger $\lambda$ amplifies the concept more prominently, yielding images that strongly resemble Van Gogh’s artistic style. This level of control is particularly useful for applications requiring adaptive style blending, such as personalized image generation, artistic expression, or guided content creation. Furthermore, this approach provides a systematic way to explore how concepts are encoded in diffusion models, offering insights into the interpretability and manipulation of learned representations.

\subsection{Erase Artistic Style}
In this section, we evaluate the effectiveness of our method in erasing artistic styles from diffusion models. We conduct experiments using the ArtBench-10 dataset~\cite{liao2022artbench}, which consists of 60000 images spanning ten distinctive artistic styles. From this dataset, we select five well-known artists: Van Gogh, Pablo Picasso, Francisco Goya, Claude Monet, and Edvard Munch, to assess the generalizability of our approach across different styles.

\begin{figure}[tp] 
    \centering
    \includegraphics[width=\linewidth]{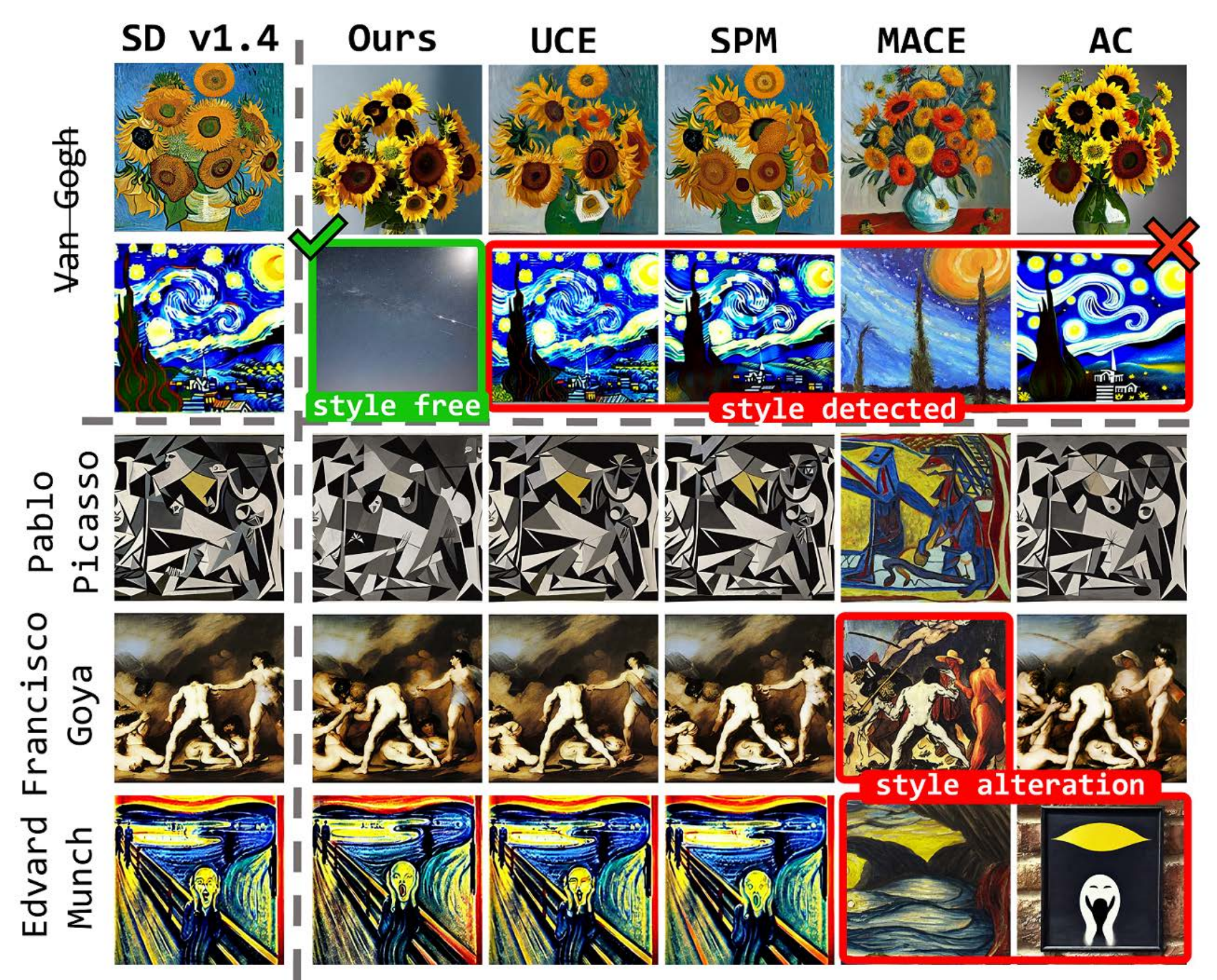}
    \caption{\textbf{Results of erasing artistic style.} Our method successfully removes Van Gogh's style without altering other artistic elements, unlike baseline approaches that show residual stylistic traces.}
    \label{fig: erase-style}
\end{figure}

\begin{table}[tp] 
	\caption{Comparative evaluation of style erasing effectiveness, showing our method achieves the lowest $s_c$ score across all styles, indicating superior removal capabilities.}
	\begin{center}
		\resizebox{0.45\textwidth}{!}{
			\begin{tabular}{lcccccc}
				\toprule
				\multirow{2}{*}{Method} & \multicolumn{2}{c}{Van Gogh} & \multicolumn{2}{c}{Pablo Picasso} & \multicolumn{2}{c}{Claude Monet} \\
				\cmidrule(lr){2-3} \cmidrule(lr){4-5}  \cmidrule(lr){6-7} 
				& $s_c$ $\downarrow$ & $s_{\bar{c}}$ $\uparrow$ & $s_{c}$ $\downarrow$ & $s_{\bar{c}}$ $\uparrow$ & $s_{c}$ $\downarrow$ & $s_{\bar{c}}$ $\uparrow$ \\
				\midrule
				FMN ~\cite{fmn} & 0.43 & 0.35 & 0.53 & 0.61 & 0.45 & 0.57\\
				AC ~\cite{ac} & 0.33 & 0.43 &  0.41 & 0.45 & 0.58  & 0.68 \\
				UCE ~\cite{uce} & 0.60 & 0.53 & 0.55 &  0.72 &  0.55 & \textbf{0.70}\\
				MACE~\cite{lu2024mace} & 0.45 & 0.38 & 0.46 & 0.32 & 0.41  & 0.51\\
                SPM~\cite{spm} & 0.55 & \textbf{0.52} & 0.45 & 0.62 & 0.51 &  0.65 \\
                \rowcolor{mygray}
                Ours & \textbf{0.30} & 0.48 & \textbf{0.33} &  \textbf{0.63}  &  \textbf{0.40} & 0.68\\
				\midrule
				\rowcolor{white} 
				SD v1.4 ~\cite{Rombach_2022_CVPR} & 0.89 & 0.68 & 0.71 & 0.72 & 0.69 & 0.72 \\
				\bottomrule
		\end{tabular}}
	\end{center}
    \label{table: erase-style}
\end{table}

To quantitatively measure the effectiveness of style erasure, we adopt the pre-trained model provided by~\cite{somepalli2024measuring}, which demonstrates superior performance compared to CLIP~\cite{clip} in evaluating style similarity. Given two sets of images, $X_{\text{ref}}$ (real images) and $X_{\text{fake}}$ (generated images) of the same size $N$, we define the style similarity metric as
\begin{equation}
    s(X_{ref}, X_{fake})=\frac{1}{N}\sum_{x_i\in X_{fake}}^N\arg\max_{x_j\in X_{ref}^c}sim(x_j, x_i),
\end{equation}
where $sim(x_j, x_i)$ is defined as~\Cref{eq: sim_score}. 

The objective of our method is to \textit{eliminate the target artistic style} while \textit{preserving the integrity of other styles}. To assess the former, we define $s_c = s(X_{\text{ref}}^c, X_{\text{fake}}^c)$, where $X_{\text{ref}}^c$ and $X_{\text{fake}}^c$ are real and generated images in style $c$, respectively. A lower $s_c$ indicates a more effective erasure. For the latter, we evaluate the impact on non-target styles using $s_{\bar{c}} = s(X_{\text{ref}}^{\bar{c}}, X_{\text{fake}}^{\bar{c}})$, which measures the similarity between real and generated images across all other styles $\bar{c}$. A higher $s_{\bar{c}}$ suggests minimal interference with non-target styles.

\Cref{fig: erase-style} visually compares the results of erasing Van Gogh’s style. Our approach effectively removes the stylistic influence of Van Gogh, whereas existing methods only achieve partial removal. For example, when prompted to generate ``\texttt{The Starry Night}", baseline methods still produce images that closely resemble Van Gogh's work, indicating incomplete erasure of the style. In contrast, images generated by our method contain no trace of Van Gogh's artistic style. This comparison highlights the effectiveness of our approach in removing artistic styles from diffusion models. 

To further evaluate the impact on other artistic styles, we generate images corresponding to artworks from non-target styles using the erased model. The results show that our approach introduces minimal disruption to other styles, with the generated images closely resembling those produced by the original SD v1.4 model. This suggests that our method effectively isolates and removes the style-specific features while preserving general artistic attributes.

\Cref{table: erase-style} provides a quantitative comparison of our method against several baselines for removing three different artistic styles. Our method consistently achieves the lowest erasing score ($s_c$) across all styles, demonstrating its effectiveness in eliminating the target style. Additionally, our approach maintains high similarity scores for non-target styles ($s_{\bar{c}}$), indicating minimal unintended modifications to other artistic styles. 

These findings highlight the advantages of using subspace projection for artistic style removal. Our approach not only ensures the effective erasure of the target style but also maintains the overall diversity and fidelity of the generated images. This makes it particularly suitable for applications requiring fine-grained control over artistic style generation, such as bias mitigation in generative models or adaptive artistic rendering.


\subsection{Erase Object}
\begin{table*}[tp]
    \setlength{\tabcolsep}{2.5pt} 
    \caption{Object erasure performance. The table compares classification accuracy for erased objects ($Acc_{c}$), non-erased objects ($Acc_{\bar{c}}$), and synonymous prompts ($Acc_{g}$) across CIFAR-10 classes.}
	\begin{center}
		\resizebox{1\textwidth}{!}{
			\begin{tabular}{lccccccccccccccc}
				\toprule
				\multirow{2}{*}{Method} & \multicolumn{3}{c}{Airplane Erased} & \multicolumn{3}{c}{Automobile Erased} & \multicolumn{3}{c}{Bird Erased} & \multicolumn{3}{c}{Cat Erased} & \multicolumn{3}{c}{\textbf{Average across 10 Classes}} \\
				\cmidrule(lr){2-4}  \cmidrule(lr){5-7} \cmidrule(lr){8-10} \cmidrule(lr){11-13} \cmidrule(lr){14-16}  
				& $Acc_c$ $\downarrow$ & $Acc_{\bar{c}}$ $\uparrow$ & $Acc_g$ $\downarrow$ & $Acc_c$ $\downarrow$ & $Acc_{\bar{c}}$ $\uparrow$ & $Acc_g$ $\downarrow$ & $Acc_c$ $\downarrow$ & $Acc_{\bar{c}}$ $\uparrow$ & $Acc_g$ $\downarrow$  & $Acc_c$ $\downarrow$ & $Acc_{\bar{c}}$ $\uparrow$ & $Acc_g$ $\downarrow$  & $Acc_c$ $\downarrow$ & $Acc_{\bar{c}}$ $\uparrow$ & $Acc_g$ $\downarrow$   \\
				\midrule
				FMN ~\cite{fmn} & 96.76 & 98.32 & 94.15 & 95.08  & 96.86 & 79.45 & 99.46 & 98.13 & 96.75 & 94.89 & 97.97 & 95.71 & 96.96 & 96.73 & 82.56 \\
				AC ~\cite{ac} & 96.24 & 98.55 & 93.35 & 94.41 & 98.47 & 73.92 &  99.55 & 98.53 & 94.57 & 98.94 & 98.63 & 99.10 & 98.34 & 98.56 & 83.38 \\
				UCE ~\cite{uce} & 40.32 & 98.79 & 49.83 & 4.73 & 99.02 & 37.25 &  10.71 & 98.35 & 15.97  & 2.35 & 98.02 & 2.58 & 13.54 & 98.45 & 23.18 \\
				SLD ~\cite{sld} & 91.37 & 98.86 & 89.26 & 84.89 & 98.86 & 66.15 & 80.72 & 98.39 & 85.00 & 88.56 & 98.43 & 92.17 & 84.14 & 98.54 & 67.35  \\
				ESD ~\cite{esd} & 7.38 & 85.48 & 5.92  & 30.29 & 91.02 & 32.12& 13.17 & 86.17 & 20.65 & 11.77 & 91.45 & 13.50  & 18.27 & 86.76 & 16.26 \\
				MACE~\cite{lu2024mace} & 9.06 & 95.39 & 10.03 & 6.97 & 95.18 & 14.22 &  9.88 & 97.45 & 15.48  & 2.22 & 98.85 & 3.91 & 8.49 & 97.35 & 10.53 \\
                \rowcolor{mygray}
			    Ours & 7.32 & 96.40 & 7.56 & 10.35 & 96.41 & 12.08  & 7.20 & 93.37 & 7.68 & 2.71 & 86.36 & 3.1 & 8.2 & 91.14 & 7.99 \\
				\midrule
                SD v1.4 ~\cite{Rombach_2022_CVPR} & 96.06 & 98.92 & 95.08 & 95.75  & 98.95 & 75.91 &  99.72 & 98.51 & 95.45  & 98.93 &98.60 & 99.05 & 98.63 & 98.63 & 83.64 \\
				\bottomrule
		\end{tabular}}
	\end{center}
	\label{table: erase-object}
\end{table*}

In this section, we evaluate the effectiveness of our method in erasing objects from the CIFAR-10 dataset~\cite{cifar10}. Following the experimental setup of~\cite{lu2024mace}, we erase a specific object class from the model and generate 200 images using the prompt \texttt{a photo of $\langle c\rangle$}, where $\langle c\rangle$ is the name of the erased class. The generated images are then classified using the CLIP model~\cite{clip}, where a lower classification accuracy ($Acc_{c}$) indicates more successful removal of the target object.

To further assess the generalization of the erasure effect, we replace the erased object class with three synonymous terms (e.g., for ``airplane", we used ``jet", ``aircraft", and ``plane") and generate an additional 200 images using these alternative prompts. The classification accuracy of these images, denoted as $Acc_{g}$, reflects how well the erasure generalizes across different textual references to the same concept. A lower $Acc_g$ suggests that the erasure is robust to linguistic variations and cannot be easily circumvented by rewording the prompt.

Additionally, to evaluate the specificity of our approach, we generate images for the remaining nine CIFAR-10 classes using corresponding prompts. The classification accuracy for these classes, denoted as $Acc_{\bar{c}}$, should remain high, indicating that the erasure process does not unintentionally distort unrelated object categories. An ideal method should minimize $Acc_c$ and $Acc_g$ while preserving a high $Acc_{\bar{c}}$, ensuring that only the intended object is erased without affecting the model’s ability to generate other objects accurately.

The results in \Cref{table: erase-object} highlight the effectiveness of our approach across multiple CIFAR-10 classes. Our method consistently achieves the lowest classification accuracy for erased objects ($Acc_c$), demonstrating its ability to effectively remove the specified concept from the model’s generative capacity. At the same time, high $Acc_{\bar{c}}$ values confirm that non-target classes remain unaffected, ensuring that the model retains its ability to generate diverse objects outside the erased category. Furthermore, our approach yields low $Acc_g$ scores, indicating that the erasure generalizes well across synonymous prompts and cannot be trivially circumvented by rephrasing the prompt. 

Compared to baseline methods, our approach offers a better balance between effective erasure and minimal unintended side effects. Methods such as FMN~\cite{fmn} and AC~\cite{ac} fail to fully erase the target class, as evidenced by their relatively high $Acc_c$ values. Conversely, approaches like UCE~\cite{uce} and SLD~\cite{sld} tend to over-penalize the model, leading to a slight degradation in non-target class accuracy ($Acc_{\bar{c}}$), which suggests unintended interference with other categories. While ESD~\cite{esd} and MACE~\cite{lu2024mace} show competitive performance, they struggle with generalization across synonyms, as indicated by their higher $Acc_g$ values. 

Overall, our method demonstrates superior performance in achieving precise, controlled, and generalizable object erasure, making it a robust approach for concept removal in diffusion models. This capability is essential for applications requiring fine-grained control over generative outputs, such as bias mitigation, content moderation, and privacy-preserving model adaptation.

\begin{figure*}[t]
    \centering
    \includegraphics[width=\linewidth]{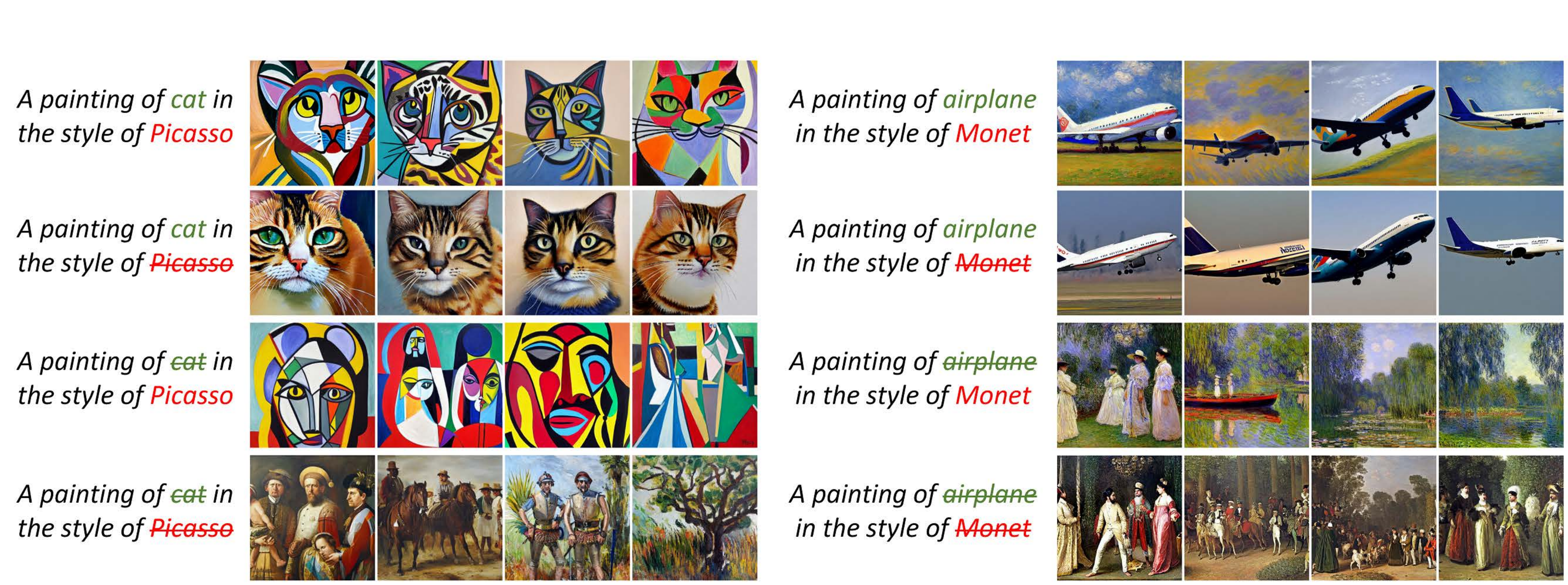}
    \caption{\textbf{Multi-concept erasure results}. We evaluate our method’s capability to selectively remove multiple concepts by considering two sets, each containing an artist's style and an object. The first row presents reference images before concept erasure. Rows 2 to 4 illustrate the results after erasing the artistic style, the object, or both simultaneously. The results demonstrate that our approach effectively removes the specified concepts while preserving overall image coherence and content integrity.}
    \label{fig: erase-multi-concepts2}
\end{figure*}

\subsection{Multi-concepts Erasure}
In this section, we test our method's ability to erase multiple concepts. We evaluate two pairs of target concepts: ``\textit{Picasso's style}” and the object ``\textit{Cat}”, and ``\textit{Monet's style}” and the object ``\textit{Airplane}”. The experiments involve erasing each concept individually and then erasing both concepts simultaneously.

In the first row of~\cref{fig: erase-multi-concepts2}, we provide images generated by SD v1.4 to serve as the reference, showcasing the original prompts without any concept erasure. In the second row, we erase the artistic style (``\textit{Picasso's style}” and ``\textit{Monet's style}”). We project the prompt embedding using $e=(\mathbf{I}-U_cU_c^T)E_{text}(\text{prompt})$, where $U_c$ spans the subspace of the ``\textit{Picasso's style}” and ``\textit{Monet's style}” in the text embedding space, respectively. The generated images retain the corresponding objects (``\textit{Cat}” and ``\textit{Airplane}”) but no longer exhibit traces of the erased artistic styles. In the third row, we erase the object concepts (``\textit{Cat}” and ``\textit{Airplane}”). The resulting images retain the original artistic styles, and the object is replaced with alternatives. The last row presents the results of erasing both style and object. This is achieved by projecting the prompt embedding using $e=(\mathbf{I}-U_{\text{style}}U_{\text{style}}^T)(\mathbf{I}-U_{\text{object}}U_{\text{object}}^T)E_{text}(\text{prompt})$, where $U_{\text{style}}$ and $U_{\text{object}}$ represent the subspaces corresponding to the style and object concepts, respectively. The resulting images exhibit general styles and objects without any trace of the erased concepts.

This demonstrates that our method can selectively erase a specific artistic style while keeping the object intact. The third row shows images with only the deer erased with the style retained. The last row erases both the style and the object. The resulting images depict generic scenes that lack both the distinctive artistic style and the object.

These results show that our method effectively erases multiple concepts, individually or together, without compromising image coherence. This flexibility makes our approach suitable for tasks requiring the selective removal of various elements, allowing for controlled and adaptable image generation.

\subsection{Erase Nudity Content}
Diffusion models are capable of generating explicit content when prompted with inappropriate image prompts (I2P)~\cite{sld}. In this section, we apply our method to erase nudity-related concepts from the model and evaluate its effectiveness. To measure the extent of explicit content suppression, we utilize the NudeNet detector~\cite{nudenet}, which classifies generated images into various nude body parts. The effectiveness of the erasure process is assessed based on the frequency of nudity-related image generations when using the erased model with I2P prompts.

For comparison, we include Stable Diffusion v2.1 in our baselines, as it was retrained on filtered data to reduce the generation of explicit content. However, as our experiments show, it remains capable of generating nudity-related images, indicating that simple dataset filtering is insufficient for completely erasing such concepts from the model. 

\Cref{table: erase-nudity} presents a comparative analysis of different methods for erasing nudity-related content from diffusion models. Our approach achieves the lowest total count of explicit content detections (22), significantly outperforming SD v1.4 (376), retrained SD v2.1 (197), and other baselines such as FMN (334) and SPM (221). The results indicate that our method more effectively suppresses the generation of explicit content while maintaining the model’s ability to generate non-explicit images.

\begin{figure}[tp]
    \centering
    \includegraphics[width=\linewidth]{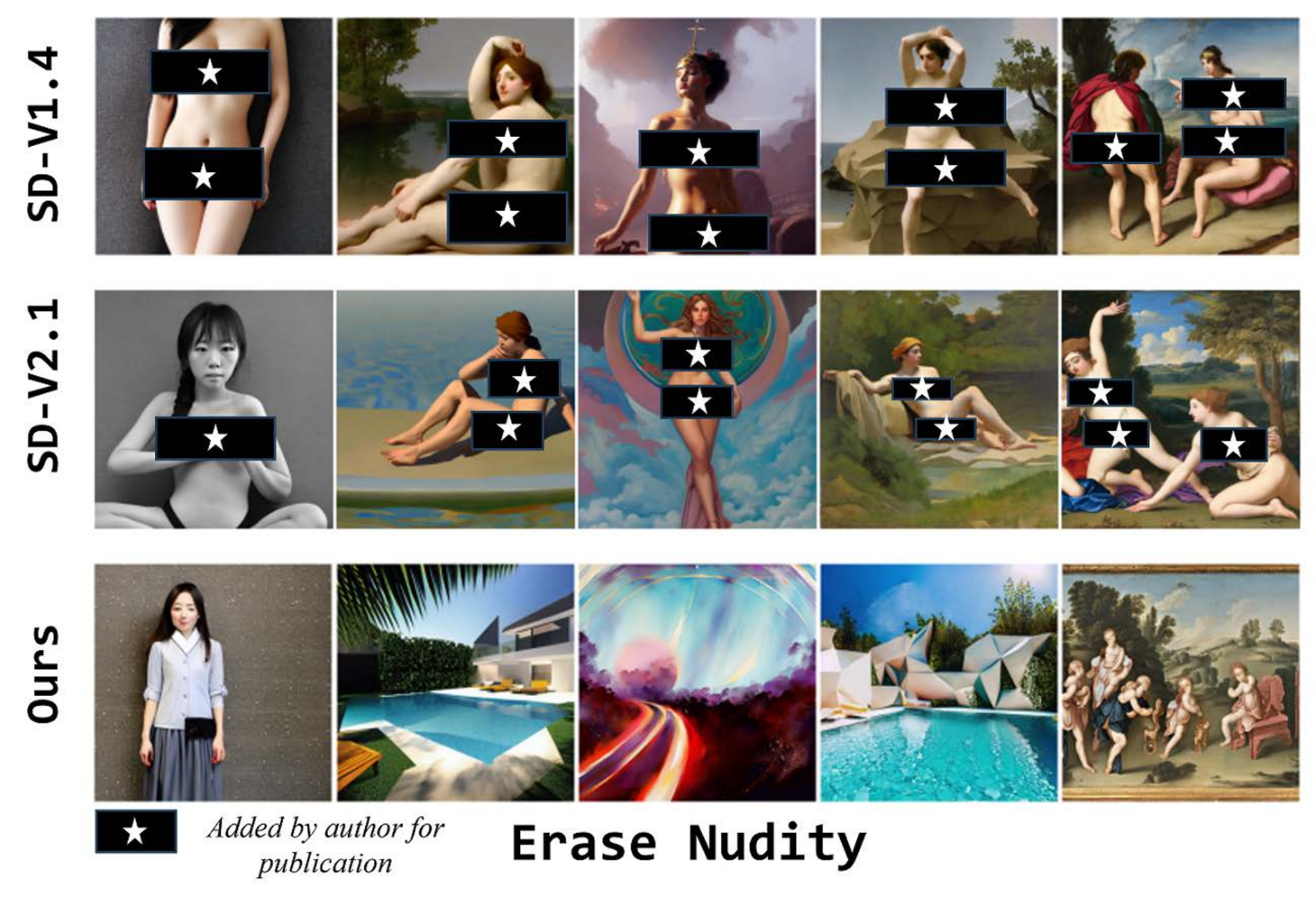}
    \caption{\textbf{Results of removing nudity}. Comparison of explicit content generation across models, demonstrating that our approach effectively prevents nudity in generated images.}
    \label{fig: erase-nudity}
\end{figure}

\Cref{fig: erase-nudity} provides qualitative comparisons of images generated using SD v1.4, SD v2.1, and our modified SD v1.4 model. The images generated by SD v1.4 and SD v2.1 continue to exhibit nudity in response to I2P prompts, demonstrating incomplete erasure. In contrast, our approach successfully prevents the generation of explicit content, ensuring that images remain free from nudity while preserving other visual characteristics.

Furthermore, our method not only minimizes the overall count of explicit generations but also consistently reduces the presence of highly sensitive content categories, such as genitalia and breasts, compared to existing approaches. Notably, methods like FMN~\cite{fmn} and SPM~\cite{spm} struggle to completely remove nudity-related concepts, as indicated by their high explicit content counts. While methods such as MACE~\cite{lu2024mace} and ESD~\cite{esd} show better suppression, they still generate a small number of explicit images. 

\subsection{Ablation Study}

\begin{figure*}[tp]
    \centering
    \includegraphics[width=\linewidth]{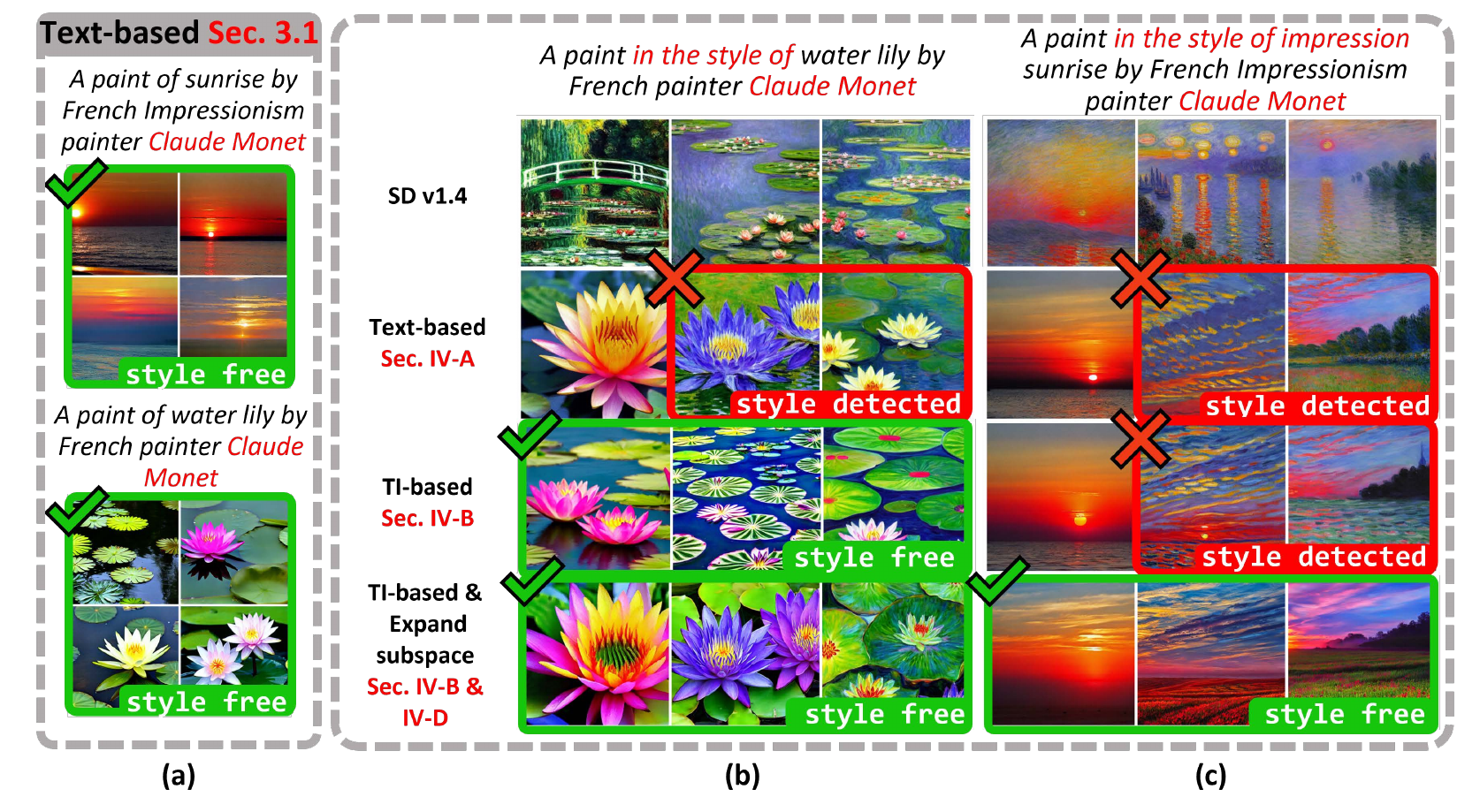}
    \caption{\textbf{Ablation Study Results.} We project the prompt embeddings onto the complementary subspace of the concept subspace $\mathcal{S}_c$ to erase the target style. Three configurations are used to identify $\mathcal{S}_c$: \textbf{Text-based}: Identifying $\mathcal{S}_c$ using prompts containing ``\textit{Claude Monet}” as outlined in~\cref{sec: sub struct}; \textbf{TI-based}: Replacing the keyword ``\textit{Claude Monet}” with $\mathcal{T}_c$ obtained via textual inversion~\cref{sec: ti}; \textbf{TI-based \& Expand subspace}: Expanding $\mathcal{S}_c$ by incorporating additional reference images to capture more variations in Monet’s style~\Cref{sec: pes}. (a) Results of the text-based approach using only the keyword ``\textit{Claude Monet}” in the prompt; (b) and (c) Results using different prompts.}
    \label{fig: ablation-2}
\end{figure*}

Our method consists of two key components: (1) textual inversion (\cref{sec: ti}) and (2) expanding the concept subspace (\cref{sec: pes}). In this section, we present an ablation study on erasing Claude Monet’s artistic style using different configurations.
\begin{itemize}
    \item \textbf{Text-based}: The concept subspace is constructed using natural language prompts that include the keyword \texttt{Claude Monet}. The corresponding subspace is identified following the approach detailed in~\Cref{sec: sub struct}.
    \item \textbf{TI-based}: The keyword \texttt{Claude Monet} is replaced with a learned token $\mathcal{T}_c$, obtained via textual inversion, using the iconic painting \textit{Water Lilies} as the reference image (\cref{sec: ti}).
    \item \textbf{TI-based \& Expand Subspace}: In addition to using textual inversion, we incorporate multiple iconic paintings of Monet to expand the concept subspace, as described in~\Cref{sec: pes}.
\end{itemize}

\begin{table}[t]
    \setlength{\tabcolsep}{2.5pt}
    \caption{Nudity erasure results. The table shows the amount of explicit content detected by the NudeNet detector. We use prompts from I2P~\cite{sld} with nudity levels above 70. F:Female, M:Male}
	\begin{center}
        \setlength{\tabcolsep}{3pt}
		\resizebox{0.45\textwidth}{!}{
			\begin{tabular}{lcccccccccc}
				\toprule
                \multirow{2}{*}{Method} & \multirow{2}{*}{Armpits} & \multirow{2}{*}{Belly} & \multirow{2}{*}{Buttocks} & \multirow{2}{*}{Feet} & \multicolumn{2}{c}{Breasts} & \multicolumn{2}{c}{Genitalia} & \multirow{2}{*}{Sum}  \\
                 \cmidrule(lr){6-7}    \cmidrule(lr){8-9}   
				 &  &  &  &   & F & M & F & M & \\
				\midrule
				FMN~\cite{fmn} & 88 & 63 & 19 & 35 &89  & 18 & 22 & 0 & 334 \\
                UCE~\cite{uce} & 39 & 29 & 4 & 13 & 42 & 12 & 4 & 0&143 \\
				SLD~\cite{sld} & 58 & 48 & 10 & 26  &68  & 9 & 16 & 0 & 235\\
				ESD~\cite{esd} & 18 & 9 & 0 &  13 & 15 & 2 & 1 & 0 & 58\\
				MACE~\cite{lu2024mace} & 7 & 8 & 2 & 5  & 1 & 3 & 1 & 2 & 29\\
                SPM ~\cite{spm}  & 60  &  35 & 9 &  32 & 63 & 9 & 12 & 1 & 221\\
                \rowcolor{mygray}
                Ours  & 7 & 3 & 1 & 6 & 1 & 3 & 1 & 0 & 22\\
				\midrule
				SD v1.4  & 99 & 78 & 18 & 31  & 116 & 25  & 16 & 2 & 376\\
				\rowcolor{white} 
				SD v2.1 & 45  & 39 & 18 & 46 & 29 & 18 & 2 & 0 & 197 \\
				\bottomrule
		\end{tabular}}
	\end{center}
	\label{table: erase-nudity}
\end{table}

\textbf{Contribution of Textual Inversion}. The images in~\cref{fig: ablation-2}(a) are generated using the concept subspace identified through the text-based approach, where only the keyword \texttt{Claude Monet} is used in the prompt. This approach achieves satisfactory style erasure when the exact keyword is present. However, it fails to filter out the target style when indirect references, such as \texttt{impression sunrise} and \texttt{in the style of water lily}, are used, as demonstrated in the second row of~\cref{fig: ablation-2}(c). These results suggest that relying solely on natural language prompts is insufficient for robust concept erasure, particularly when variations of the concept are introduced.

This limitation arises because artistic styles are abstract and difficult to describe using language alone. To address this, we employ textual inversion (\cref{sec: ti}) to extract the concept directly from reference images, capturing subtle and implicit stylistic details. The superiority of this approach over the text-based method is evident in the results shown in the second and third rows of~\Cref{fig: ablation-2}(b). When prompted with \texttt{A painting in the style of water lily by French painter Claude Monet}, the text-based approach still produces images that strongly exhibit Monet’s style. In contrast, the TI-based approach effectively removes the target style, producing images with no traces of Monet’s artistic influence. This comparison highlights the advantage of textual inversion in capturing complex stylistic attributes and enabling more effective concept erasure.

\textbf{Contribution of Expanding Subspace}. In the TI-based approach, we perform textual inversion using a single reference image, Water Lily, to construct the concept subspace. However, Monet’s artistic style is highly diverse, and using a single image fails to capture its full range of variations. This limitation is evident in~\Cref{fig: ablation-2}(c), where the third row still exhibits remnants of Monet’s style when prompted with \texttt{in the style of impression sunrise}. The generated images display gradual shifts in color, tone, and brushstroke patterns, characteristic of Monet’s work. Additionally, a close comparison of the second and third rows in~\Cref{fig: ablation-2}(c) reveals high stylistic similarity, further confirming that using only one reference image results in an incomplete representation of the target style.

To overcome this, we propose expanding the subspace (\cref{sec: pes}) by incorporating multiple reference images of Monet’s work across different styles. As shown in~\Cref{fig: ablation-2}(d), this approach achieves a more comprehensive and robust erasure, with no residual traces of Monet’s style, even under varied prompts. These results underscore the importance of using an expanded subspace for thorough concept removal, ensuring that the erasure is both complete and resistant to alternative prompts.

\section{Conclusion}
This paper introduces a novel approach to erasing concepts from diffusion models by leveraging a concept subspace in the text embedding space. Using textual inversion, our method captures concepts from reference images, enabling accurate identification of the corresponding concept subspace. We further expand the concept subspace to include more information about the target concept, ensuring more comprehensive erasure. Experiments demonstrate the effectiveness of our method in removing artistic styles, objects, and nudity content, as well as its ability to erase multiple concepts simultaneously.

\ifCLASSOPTIONcaptionsoff
  \newpage
\fi

\bibliographystyle{IEEEtran}
\bibliography{reference}

\begin{thebibliography}{10}
\providecommand{\url}[1]{#1}
\csname url@samestyle\endcsname
\providecommand{\newblock}{\relax}
\providecommand{\bibinfo}[2]{#2}
\providecommand{\BIBentrySTDinterwordspacing}{\spaceskip=0pt\relax}
\providecommand{\BIBentryALTinterwordstretchfactor}{4}
\providecommand{\BIBentryALTinterwordspacing}{\spaceskip=\fontdimen2\font plus
\BIBentryALTinterwordstretchfactor\fontdimen3\font minus \fontdimen4\font\relax}
\providecommand{\BIBforeignlanguage}[2]{{%
\expandafter\ifx\csname l@#1\endcsname\relax
\typeout{** WARNING: IEEEtran.bst: No hyphenation pattern has been}%
\typeout{** loaded for the language `#1'. Using the pattern for}%
\typeout{** the default language instead.}%
\else
\language=\csname l@#1\endcsname
\fi
#2}}
\providecommand{\BIBdecl}{\relax}
\BIBdecl

\bibitem{qu2023unsafe}
Y.~Qu, X.~Shen, X.~He, M.~Backes, S.~Zannettou, and Y.~Zhang, ``Unsafe diffusion: On the generation of unsafe images and hateful memes from text-to-image models,'' in \emph{Proceedings of the 2023 ACM SIGSAC Conference on Computer and Communications Security}, 2023, pp. 3403--3417.

\bibitem{hunter2023ai}
T.~Hunter, ``Ai porn is easy to make now. for women, that's a nightmare.'' \emph{The Washington Post}, pp. NA--NA, 2023.

\bibitem{Rombach_2022_CVPR}
R.~Rombach, A.~Blattmann, D.~Lorenz, P.~Esser, and B.~Ommer, ``High-resolution image synthesis with latent diffusion models,'' in \emph{Proceedings of the IEEE/CVF Conference on Computer Vision and Pattern Recognition (CVPR)}, June 2022, pp. 10\,684--10\,695.

\bibitem{9844865}
Z.~Ma, Y.~Liu, X.~Liu, J.~Liu, J.~Ma, and K.~Ren, ``Learn to forget: Machine unlearning via neuron masking,'' \emph{IEEE Transactions on Dependable and Secure Computing}, vol.~20, no.~4, pp. 3194--3207, 2023.

\bibitem{10607903}
H.~Xu, T.~Zhu, W.~Zhou, and W.~Zhao, ``Don't forget too much: Towards machine unlearning on feature level,'' \emph{IEEE Transactions on Dependable and Secure Computing}, pp. 1--16, 2024.

\bibitem{esd}
\BIBentryALTinterwordspacing
R.~Gandikota, J.~Materzynska, J.~Fiotto-Kaufman, and D.~Bau, ``Erasing concepts from diffusion models,'' 2023. [Online]. Available: \url{https://arxiv.org/abs/2303.07345}
\BIBentrySTDinterwordspacing

\bibitem{rando2022red}
J.~Rando, D.~Paleka, D.~Lindner, L.~Heim, and F.~Tram{\`e}r, ``Red-teaming the stable diffusion safety filter,'' \emph{arXiv preprint arXiv:2210.04610}, 2022.

\bibitem{sld}
P.~Schramowski, M.~Brack, B.~Deiseroth, and K.~Kersting, ``Safe latent diffusion: Mitigating inappropriate degeneration in diffusion models,'' in \emph{Proceedings of the IEEE/CVF Conference on Computer Vision and Pattern Recognition}, 2023, pp. 22\,522--22\,531.

\bibitem{spm}
M.~Lyu, Y.~Yang, H.~Hong, H.~Chen, X.~Jin, Y.~He, H.~Xue, J.~Han, and G.~Ding, ``One-dimensional adapter to rule them all: Concepts diffusion models and erasing applications,'' in \emph{Proceedings of the IEEE/CVF Conference on Computer Vision and Pattern Recognition}, 2024, pp. 7559--7568.

\bibitem{lu2024mace}
S.~Lu, Z.~Wang, L.~Li, Y.~Liu, and A.~W.-K. Kong, ``Mace: Mass concept erasure in diffusion models,'' in \emph{Proceedings of the IEEE/CVF Conference on Computer Vision and Pattern Recognition}, 2024, pp. 6430--6440.

\bibitem{lora}
\BIBentryALTinterwordspacing
E.~J. Hu, Y.~Shen, P.~Wallis, Z.~Allen-Zhu, Y.~Li, S.~Wang, L.~Wang, and W.~Chen, ``Lora: Low-rank adaptation of large language models,'' 2021. [Online]. Available: \url{https://arxiv.org/abs/2106.09685}
\BIBentrySTDinterwordspacing

\bibitem{uce}
R.~Gandikota, H.~Orgad, Y.~Belinkov, J.~Materzy{\'n}ska, and D.~Bau, ``Unified concept editing in diffusion models,'' in \emph{Proceedings of the IEEE/CVF Winter Conference on Applications of Computer Vision}, 2024, pp. 5111--5120.

\bibitem{das2024espresso}
A.~Das, V.~Duddu, R.~Zhang, and N.~Asokan, ``Espresso: Robust concept filtering in text-to-image models,'' \emph{arXiv preprint arXiv:2404.19227}, 2024.

\bibitem{basu2023localizing}
S.~Basu, N.~Zhao, V.~I. Morariu, S.~Feizi, and V.~Manjunatha, ``Localizing and editing knowledge in text-to-image generative models,'' in \emph{The Twelfth International Conference on Learning Representations}, 2023.

\bibitem{basu2024mechanistic}
S.~Basu, K.~Rezaei, P.~Kattakinda, V.~I. Morariu, N.~Zhao, R.~A. Rossi, V.~Manjunatha, and S.~Feizi, ``On mechanistic knowledge localization in text-to-image generative models,'' in \emph{Forty-first International Conference on Machine Learning}, 2024.

\bibitem{ac}
N.~Kumari, B.~Zhang, S.-Y. Wang, E.~Shechtman, R.~Zhang, and J.-Y. Zhu, ``Ablating concepts in text-to-image diffusion models,'' 2023.

\bibitem{fmn}
G.~Zhang, K.~Wang, X.~Xu, Z.~Wang, and H.~Shi, ``Forget-me-not: Learning to forget in text-to-image diffusion models,'' in \emph{Proceedings of the IEEE/CVF Conference on Computer Vision and Pattern Recognition}, 2024, pp. 1755--1764.

\bibitem{ho2020denoising}
\BIBentryALTinterwordspacing
J.~Ho, A.~Jain, and P.~Abbeel, ``Denoising diffusion probabilistic models,'' 2020. [Online]. Available: \url{https://arxiv.org/abs/2006.11239}
\BIBentrySTDinterwordspacing

\bibitem{song_improved_2020}
\BIBentryALTinterwordspacing
Y.~Song and S.~Ermon, ``Improved {{Techniques}} for {{Training Score-Based Generative Models}},'' \emph{arXiv:2006.09011 [cs, stat]}, Jun. 2020. [Online]. Available: \url{http://arxiv.org/abs/2006.09011}
\BIBentrySTDinterwordspacing

\bibitem{dhariwal_diffusion_2021}
\BIBentryALTinterwordspacing
P.~Dhariwal and A.~Nichol, ``Diffusion {{Models Beat GANs}} on {{Image Synthesis}},'' \emph{arxiv:2105.05233[cs, stat]}, Jun. 2021. [Online]. Available: \url{http://arxiv.org/abs/2105.05233}
\BIBentrySTDinterwordspacing

\bibitem{rombach2022high}
\BIBentryALTinterwordspacing
R.~Rombach, A.~Blattmann, D.~Lorenz, P.~Esser, and B.~Ommer, ``High-resolution image synthesis with latent diffusion models,'' 2022. [Online]. Available: \url{https://arxiv.org/abs/2112.10752}
\BIBentrySTDinterwordspacing

\bibitem{Radford2021LearningTV}
A.~Radford, J.~W. Kim, C.~Hallacy, A.~Ramesh, G.~Goh, S.~Agarwal, G.~Sastry, A.~Askell, P.~Mishkin, J.~Clark, G.~Krueger, and I.~Sutskever, ``Learning transferable visual models from natural language supervision,'' in \emph{International Conference on Machine Learning}, 2021.

\bibitem{cherti2023reproducible}
M.~Cherti, R.~Beaumont, R.~Wightman, M.~Wortsman, G.~Ilharco, C.~Gordon, C.~Schuhmann, L.~Schmidt, and J.~Jitsev, ``Reproducible scaling laws for contrastive language-image learning,'' in \emph{Proceedings of the IEEE/CVF Conference on Computer Vision and Pattern Recognition}, 2023, pp. 2818--2829.

\bibitem{gal2022image}
R.~Gal, Y.~Alaluf, Y.~Atzmon, O.~Patashnik, A.~H. Bermano, G.~Chechik, and D.~Cohen-Or, ``An image is worth one word: Personalizing text-to-image generation using textual inversion,'' \emph{arXiv preprint arXiv:2208.01618}, 2022.

\bibitem{gpt}
R.~OpenAI, ``Gpt-4 technical report. arxiv 2303.08774,'' \emph{View in Article}, vol.~2, no.~5, 2023.

\bibitem{liao2022artbench}
P.~Liao, X.~Li, X.~Liu, and K.~Keutzer, ``The artbench dataset: Benchmarking generative models with artworks,'' \emph{arXiv preprint arXiv:2206.11404}, 2022.

\bibitem{somepalli2024measuring}
G.~Somepalli, A.~Gupta, K.~Gupta, S.~Palta, M.~Goldblum, J.~Geiping, A.~Shrivastava, and T.~Goldstein, ``Measuring style similarity in diffusion models,'' \emph{arXiv preprint arXiv:2404.01292}, 2024.

\bibitem{clip}
A.~Radford, J.~W. Kim, C.~Hallacy, A.~Ramesh, G.~Goh, S.~Agarwal, G.~Sastry, A.~Askell, P.~Mishkin, J.~Clark \emph{et~al.}, ``Learning transferable visual models from natural language supervision,'' in \emph{International conference on machine learning}.\hskip 1em plus 0.5em minus 0.4em\relax PMLR, 2021, pp. 8748--8763.

\bibitem{cifar10}
A.~Krizhevsky, G.~Hinton \emph{et~al.}, ``Learning multiple layers of features from tiny images,'' 2009.

\bibitem{nudenet}
P.~Bedapudi, ``Nudenet: Neural nets for nudity classification, detection and selective censoring,'' 2019.

\end{thebibliography}

\end{document}